\documentclass[11pt]{article}

\usepackage{acl}

\usepackage{times}
\usepackage{latexsym}

\usepackage[T1]{fontenc}
\usepackage[utf8]{inputenc}

\usepackage{microtype}

\usepackage{inconsolata}

\usepackage{graphicx}
\usepackage{tabularx, multirow, booktabs}
\usepackage{xcolor}      % 颜色支持
\usepackage{colortbl}    % 表格着色支持

\usepackage{longtable}
\usepackage{booktabs}
\usepackage{multirow}
\usepackage{array}
\usepackage{arydshln}
\usepackage{graphicx}
\usepackage{amsmath}
\usepackage{amssymb}
\usepackage{pifont}
\usepackage{arydshln}

\usepackage{microtype}
\usepackage[utf8]{inputenc}
\usepackage{latexsym}
\usepackage{makecell}
\usepackage[ruled,vlined]{algorithm2e}
\usepackage{enumitem}
\usepackage{caption}
\usepackage{listings}
\usepackage{fvextra}
\usepackage{algpseudocode}
\usepackage{booktabs,makecell,siunitx,adjustbox,arydshln}
\usepackage{hyperref}
\usepackage[most]{tcolorbox}
\tcbuselibrary{breakable, listings, skins}
\usetikzlibrary{positioning, arrows.meta}

\definecolor{citecolor}{HTML}{0071bc}
\hypersetup{
    colorlinks=true,
    linkcolor=red,
    anchorcolor=citecolor,
    citecolor=citecolor
}
\definecolor{lightbluebg}{RGB}{235, 245, 255}  % 一个非常淡的背景蓝
\definecolor{blueframe}{RGB}{70, 130, 180}     % 钢蓝色边框 (SteelBlue)

\usepackage[title,titletoc]{appendix}
\setlist[enumerate]{leftmargin=*, label=(\arabic*), topsep=0pt}
\setlist[itemize]{leftmargin=*, topsep=0pt}
\newcommand{\stddev}[1]{\textcolor{gray}{\scalebox{.8}{$\pm$#1}}}

\newcommand{\cmark}{\textcolor{green!50!black}{\checkmark}}   % Green checkmark
\newcommand{\xmark}{\textcolor{red}{\ding{55}}}
\definecolor{yellowtext}{RGB}{68,132,243}
\definecolor{yellowred}{RGB}{50,167,82}
\definecolor{yellowblue}{RGB}{251,191,5}
\newcommand{\TextCircle}[1][0.7]{%
    \tikz[baseline=(char.base)]\node[shape=circle,draw=black,inner sep=1.5pt,line width=0.5pt,fill=yellowtext,text=white,scale=#1] (char) {T};\hspace{-1pt}
}
\newcommand{\ImageCircle}[1][0.76]{%
    \tikz[baseline=(char.base)]\node[shape=circle,draw=black,inner sep=1.5pt,line width=0.5pt,fill=yellowred,text=white,scale=#1] (char) {I};\hspace{-1pt}
}

\definecolor{darkblue}{rgb}{0, 0, 0.5}
\hypersetup{colorlinks=true, citecolor=darkblue, linkcolor=darkblue, urlcolor=darkblue}

\title{MedMMV: Evidence-Grounded Trajectory Control for More Reliable Multimodal Clinical Reasoning}

\author{%
 \textbf{Hongjun Liu}$^{1}$,
 \textbf{Yinghao Zhu}$^{2}$,  
 \textbf{Yuhui Wang}$^{3}$,
 \textbf{Yitao Long}$^{1}$,
 \textbf{Lequan Yu}$^{2}$, 
 \textbf{Chen Zhao}$^{1}$
   \vspace{.5em} 
  \\
  $^1$New York University, 
  $^2$The University of Hong Kong,
  $^3$Independent Researcher, \\
}
\begin{document}
\maketitle
\begin{abstract}
Current multimodal large language models (MLLMs) suffer from unstable reasoning due to the lack of mechanisms for controlling reasoning trajectories, often leading to hallucinated conclusions. Our analysis reveals that early-stage uncertainty induces trajectory switching, significantly increasing hallucination risk.
We propose MedMMV, an evidence-grounded trajectory control framework that treats reasoning as a search-and-selection process over candidate trajectories. By constructing, refining, and selecting among multiple reasoning paths, MedMMV improves reliability and produces structured supervision signals over trajectory quality.
We further convert these signals into learnable objectives for post-training, including reject sampling and pairwise preference learning, enabling smaller models to learn trajectory selection as an efficient single-pass capability. Instead of merely imitating reasoning traces, the model is trained to assign higher likelihood to preferred trajectories.
Experiments show that richer supervision leads to stronger transfer, with accuracy improving from 10.5\% to 29.0\%. Our results demonstrate that trajectory selection is the key transferable capability for reliable reasoning.
\end{abstract}

\section{Introduction}

\begin{figure*}[!t]
\begin{center}
% \vspace{-10mm}
 % \includegraphics[width = \linewidth]{figures/halluc_case_performance_new.pdf}
 \includegraphics[width = \linewidth]{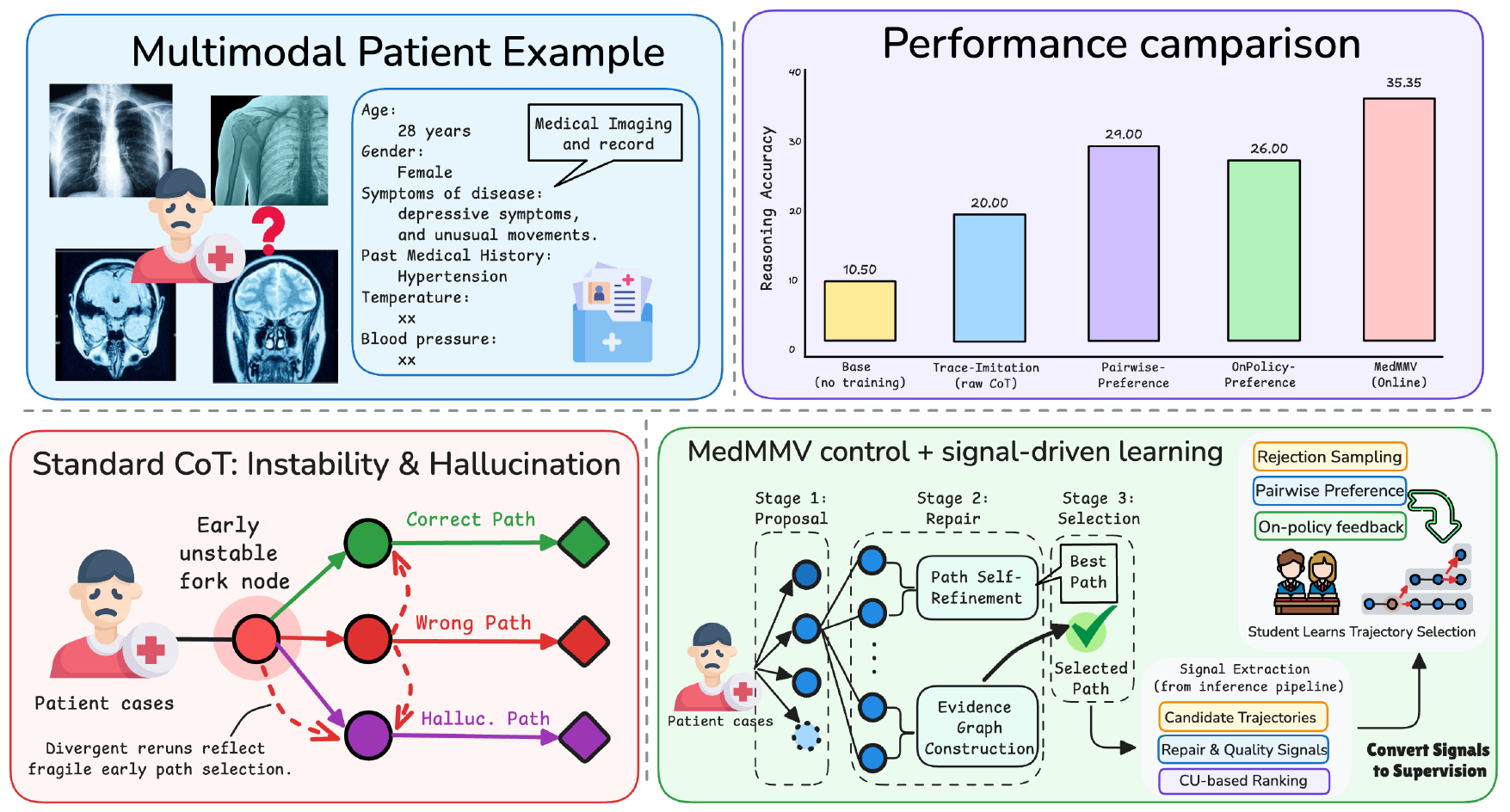}
\end{center}
\vspace{-4mm}
\caption{
% Overview of MedMMV as a trajectory-level control framework for multimodal clinical reasoning. \emph{Top-left}: A representative multimodal patient case combining medical images and structured clinical context. \emph{Top-right}: Performance comparison on medical benchmarks, showing consistent gains over standard CoT and strong recovery by the on-policy student. \emph{Bottom-left}: Standard CoT is fragile because early trajectory instability can lead to divergent, incorrect, or hallucinated reasoning paths. \emph{Bottom-right}: MedMMV improves reliability through trajectory-level control, and these same control signals can be partially amortized into a smaller student model through on-policy distillation.
Overview of MedMMV for trajectory-level control and signal-driven learning in multimodal clinical reasoning.
\emph{Top-left}: A multimodal patient case.
\emph{Top-right}: The student model trained with trajectory-level supervision significantly outperforms trace imitation and approaches the performance of full trajectory control.
\emph{Bottom-left}: Standard chain-of-thought reasoning suffers from early uncertainty, leading to trajectory divergence.
\emph{Bottom-middle}: MedMMV performs trajectory-level control through three stages.
\emph{Bottom-right}: The MedMMV process produces structured trajectory-level signals, which we transfer to a smaller model through post-training.
}

\vspace{-4mm}
\label{fig:illustrate_fig}
\end{figure*}

Recent multimodal large language models (MLLMs)~\citep{anthropic_claude4_sonnet, openai2025gpt5}, have demonstrated strong reasoning capabilities in clinical settings, approaching expert-level performance on medical question answering benchmarks~\citep{Singhal2025, Bhayana2023, Sandmann2025,bedi2025medhelmholisticevaluationlarge}. 
Despite this progress, deploying agent-based systems in real-world healthcare is challenging due to the need for extreme reliability~\citep{zhu2025medagentboard}. Shown in Figure~\ref{fig:illustrate_fig}, existing systems face the primary issue of their inability to reliably select among competing reasoning trajectories. In practice, small variations in early-stage reasoning can lead to divergent trajectories, resulting in unstable decisions and hallucinated conclusions. Our pilot analysis reveals a structured failure pattern: early-stage uncertainty induces trajectory switching, increasing hallucination risk by 13\%. 
% These findings suggest that reliability in clinical reasoning requires explicit control over reasoning trajectories, rather than relying solely on single-path generation.

To solve this problem, we introduce MedMMV, a trajectory-level control framework that reformulates multimodal reasoning as a search-and-selection process over candidate trajectories. Instead of committing to a single chain of thought, MedMMV explicitly explores, refines, and evaluates multiple reasoning paths through three primitives: (1) proposal of diverse candidate trajectories, (2) evidence-grounded repair that enforces factual consistency via a multimodal evidence graph, and (3) selection of the most reliable trajectory using Combined Uncertainty (CU) score. This formulation elevates reasoning from single-path generation to trajectory-level decision-making, enabling the model to explicitly compare and select among competing hypotheses under uncertainty.

Beyond improving reasoning reliability, MedMMV produces structured trajectory-level supervision signals, including candidate proposals, step-level repair feedback, and CU-based trajectory preferences. We convert these signals into learnable objectives by constructing supervision over trajectory quality, including rejection sampling and pairwise preference learning, which bias the model toward high-quality reasoning paths. This yields a key insight: trajectory control frameworks can serve as supervision generators, rather than merely inference procedures. Therefore, we transfer these signals to a smaller model through post-training, enabling it to \emph{amortize} trajectory selection into a single-pass generation process. The student model is trained to assign higher likelihood to preferred trajectories, allowing it to implicitly perform trajectory selection at inference time without explicit multi-path search.

Empirically, MedMMV demonstrates consistent gains across both regimes. At inference time, it improves performance on six public medical benchmarks, achieving gains of 7.9\% on MedXpert-MM and 12.7\% on MedFrameQA with GPT-5 as the executor, while also enhancing reliability metrics. Physician studies further confirm improved truthfulness. In the amortized regime, progressively richer supervision signals lead to stronger transfer performance: accuracy improves from 10.5\% with trace imitation to 29.0\% with trajectory-level preference learning. These results highlight that learning to distinguish and select among competing trajectories, rather than merely imitating reasoning traces, is the key transferable capability for reliable reasoning.

In summary, our contributions are threefold:
(1) \textbf{Trajectory-level failure analysis.} We identify a structured failure mode in multimodal clinical reasoning, where early-stage uncertainty leads to trajectory switching and increased hallucination.
(2) \textbf{Trajectory control and amortized selection.} We propose MedMMV, an evidence-grounded framework, and show how its trajectory-level signals can be converted into learnable objectives (e.g., rejection sampling) and transferred to smaller models for efficient single-pass trajectory selection.
(3) \textbf{Comprehensive evaluation.} We demonstrate the effectiveness of our approach across multiple settings, validating both reliability gains and efficient capability transfer.

\section{Related Work}

\textbf{Reliability, consistency, and hallucination in multimodal clinical reasoning.}
Ensuring evidence-faithful reasoning is especially challenging in high-stakes multimodal clinical settings, where noisy or misaligned evidence can induce random guessing and cross-modal hallucinations~\citep{liu2024surveyhallucinationlargevisionlanguage,bai2025hallucinationmultimodallargelanguage}. Medical MLLMs such as LLaVA-Med~\citep{li2023llavamedtraininglargelanguageandvision} and Med-Flamingo~\citep{moor2023medflamingomultimodalmedicalfewshot} have advanced image-text medical QA, but they still lack strong uncertainty calibration and step-level verifiability. Related benchmarks and analyses, including POPE~\citep{li2023evaluatingobjecthallucinationlarge}, HallusionBench~\citep{guan2024hallusionbenchadvanceddiagnosticsuite}, Med-HALT~\citep{pal2023medhaltmedicaldomainhallucination}, and ConBench~\citep{zhang2024unveilingtapestryconsistencylarge}, expose hallucination patterns and consistency failures, while prior work also documents the ``snowballing'' dynamics of multimodal hallucination~\citep{zhong2024investigatingmitigatingmultimodalhallucination}. 

\begin{table*}[!t]
\centering
\vspace{-4mm}
\resizebox{\textwidth}{!}{%
\begin{tabular}{@{}l cc  ccc  cccc @{}}
\toprule
\multirow{3}{*}{\textbf{System}} & 
\multicolumn{2}{c}{\textbf{Data Modality}} & 
\multicolumn{3}{c}{\textbf{Core Reasoning Engine}} & 
\multicolumn{4}{c}{\textbf{Structured Reasoning \& Validation}} \\
\cmidrule(lr){2-3} \cmidrule(lr){4-6} \cmidrule(lr){7-10}
& \multirow{2}{*}{Text} & \multirow{2}{*}{Image} & 
\multirow{2}{*}{\begin{tabular}[c]{@{}c@{}}Self- \\ Revision\end{tabular}} & 
\multirow{2}{*}{\begin{tabular}[c]{@{}c@{}}Evidence \\ Grounding\end{tabular}} & 
\multirow{2}{*}{\begin{tabular}[c]{@{}c@{}}Halluc. \\ Control\end{tabular}} & 
\multirow{2}{*}{\begin{tabular}[c]{@{}c@{}}Parallel \\ Exploration\end{tabular}} & 
\multirow{2}{*}{\begin{tabular}[c]{@{}c@{}}Active \\ Supervision\end{tabular}} & 
\multirow{2}{*}{\begin{tabular}[c]{@{}c@{}}Consensus \& \\ Aggregation\end{tabular}} & 
\multirow{2}{*}{\begin{tabular}[c]{@{}c@{}}Traceable \\ Reasoning Path\end{tabular}} \\
& & & & & & & & & \\
\midrule
\multicolumn{10}{@{}l}{\textit{Reasoning Methods}} \\
\hdashline
CoT~\citep{wei2023chainofthoughtpromptingelicitsreasoning} & \cmark & \xmark & \xmark & \xmark & \xmark & \xmark & \xmark & \xmark & \xmark \\
MedPaLM~\citep{singhal2023expertlevelmedicalquestionanswering} & \cmark & \cmark$^*$ & \cmark & \cmark & \cmark & \xmark & \xmark & \cmark$^{**}$ & \xmark \\
\midrule
\multicolumn{10}{@{}l}{\textit{Multi-Agent Systems}} \\
\hdashline
ClinicalAgent~\citep{yue2024clinicalagentclinicaltrialmultiagent} & \cmark & \xmark & \cmark & \xmark & \xmark & \xmark & \xmark & \xmark & \xmark \\
MDAgents~\citep{kim2024mdagentsadaptivecollaborationllms} & \cmark & \cmark & \cmark & \xmark & \xmark & \xmark & (\cmark) & (\cmark) & \xmark \\
ColaCare~\citep{wang2025colacare} & \cmark & \xmark & \cmark & \cmark & \xmark & \xmark & (\cmark) & \cmark & \xmark \\
ReConcile~\citep{chen2024reconcileroundtableconferenceimproves} & \cmark & \xmark & \cmark & \xmark & \xmark & \xmark & (\cmark) & \cmark & \xmark \\
\midrule
\multicolumn{10}{@{}l}{\textit{Agentic Workflow Automation}} \\
\hdashline
AFlow~\citep{zhang2025aflowautomatingagenticworkflow} & \cmark & \xmark & \cmark & \xmark & \xmark & \xmark & \xmark & \xmark & \xmark \\
ADAS~\citep{hu2025automateddesignagenticsystems} & \cmark & \xmark & \cmark & \xmark & \xmark & \xmark & \xmark & \xmark & \xmark \\
\midrule
\textbf{MedMMV (ours)} & \cmark & \cmark & \cmark & \cmark & \cmark & \cmark & \cmark & \cmark & \cmark \\
\bottomrule
\multicolumn{10}{l}{\footnotesize{$^*$MedPaLM-M, a variant, handles multiple modalities. $^{**}$Achieved via self-consistency. (\cmark) Denotes partial or implicit support.}}
\end{tabular}%
}
\vspace{-3mm}
\caption{Comparison of MedMMV with representative prior methods.}
\label{tab:related_work_comparison}
\vspace{-4mm}
\end{table*}
\noindent\textbf{Controllable reasoning and process-level regulation.}
A growing line of work argues that supervising reasoning processes is more reliable than scoring final outcomes alone~\citep{lightman2023letsverifystepstep,stiennon2022learningsummarizehumanfeedback}. In parallel, uncertainty estimation has been used to identify when model outputs should be trusted or deferred~\citep{kadavath2022languagemodelsmostlyknow}. Inference-time exploration methods such as self-consistency~\citep{wang2023selfconsistencyimproveschainthought}, Tree-of-Thoughts~\citep{yao2023treethoughtsdeliberateproblem}, and multi-agent reasoning or debate frameworks~\citep{kim2024mdagentsadaptivecollaborationllms,chen2024reconcileroundtableconferenceimproves,li2025swarmsysdecentralizedswarminspiredagents} improve reasoning diversity.
% , but they often remain vulnerable to correlated errors and lack explicit grounding constraints. 

\noindent\textbf{Distillation and training-time transfer of reasoning control.}
% While inference-time control can improve reliability, its computational cost motivates training-time alternatives. 
In medicine, knowledge distillation from stronger teacher models has been used to improve medical QA systems~\citep{singhal2023expertlevelmedicalquestionanswering}. More broadly, rejection sampling and process-level rewards have shown promise for transferring stronger reasoning behaviors~\citep{lightman2023letsverifystepstep}, and on-policy distillation has emerged as a way to transfer search-based or feedback-driven reasoning into single-pass models by reranking student-generated rollouts~\citep{stiennon2022learningsummarizehumanfeedback}

Unlike prior work, we connect trajectory-level control to efficient deployment. We combine parallel trajectory exploration, evidence-grounded repair, and uncertainty-aware selection into a unified framework and further incorporate stage-wise feedback into post-training, teaching smaller models to select more reliable reasoning trajectories in a single pass.

\section{Pilot Analysis}
\label{sec:preliminary}

We conduct a pilot audit on medical QA and VQA to probe instability in multi-step reasoning. To quantify this phenomenon, we introduce two complementary diagnostics across repeated generations of the same input: (i) the \emph{random guess measure} (RGM), which captures rollout-to-rollout dispersion in intermediate decisions via entropy over sampled choices, and (ii) the \emph{cross-modal hallucination rate} (CMHR), which evaluates answer reliability by combining truthfulness, informativeness, and visual consistency scores.
% Intuitively, RGM reflects how unstable the reasoning trajectory is, while CMHR measures whether the resulting outputs remain grounded.
As shown in Figure~\ref{fig:pilot-case}, we observe a consistent pattern: \emph{instability, path switching, and hallucination are tightly coupled along reasoning trajectories}. Higher prior entropy (RGM) predicts elevated hallucination risk in subsequent steps, and majority-path switching events coincide with sharp increases in high-CMHR probability. At the trajectory level, questions with frequent switching exhibit significantly higher average hallucination rates, indicating that local instability accumulates into global inconsistency, which is further suggested by an exploratory mediation analysis. For detailed case studies, complementary diagnostics definition and derivation details, please refer to Appendix~\ref{app:pilot-metrics}. 

% , revealing a structured failure mode we term \emph{trajectory drift}.

\noindent\textbf{Takeaway.} 
These findings suggest that hallucination is not merely a terminal error but an emergent property of unstable reasoning dynamics. This motivates our framework to intervene directly on reasoning trajectories via structured proposal, repair, and selection. Detailed definitions, statistical analyses, and robustness checks are provided in Appendix~\ref{app:analysis-details}.

\section{Methodology}
Motivated by a pilot audit, we formulate multimodal clinical reasoning as a \emph{trajectory selection problem} and introduce MedMMV. 
Given clinical notes $\mathcal{T}$ and images $\mathcal{I}$, MedMMV produces a diagnosis $\hat{y}$, a reasoning trajectory $\hat{p}$, and a supporting evidence graph $\mathcal{E}_{\hat{p}}$. 
To ensure reliable results, MedMMV models reasoning as a \emph{search-and-selection process} across three stages, \emph{proposal}, \emph{repair}, and \emph{selection}, to construct, refine, and evaluate candidate paths.

\begin{figure*}[!ht]
\centering
\vspace{-2mm}
\includegraphics[width = \linewidth]{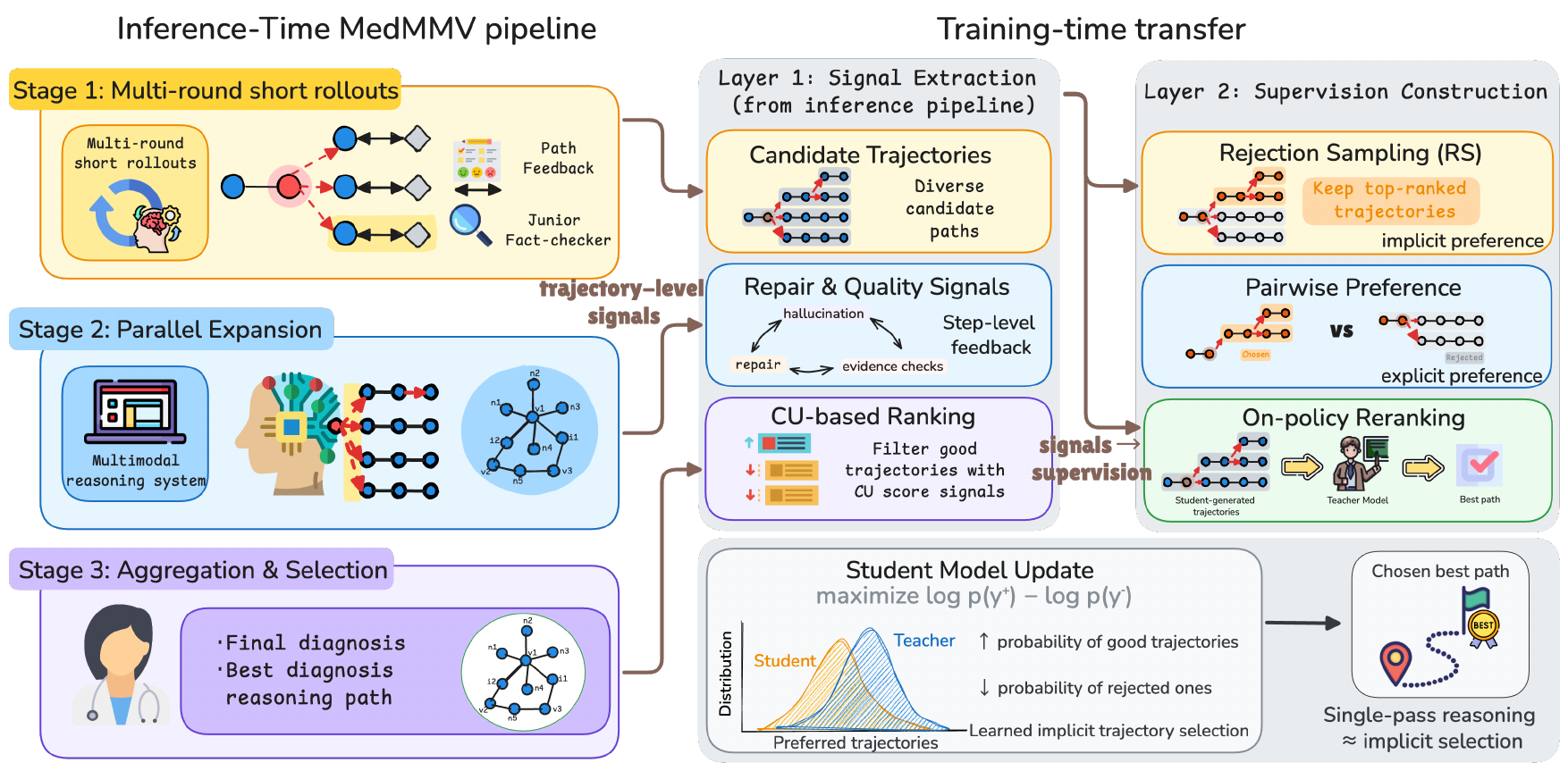}
\vspace{-7mm}
\caption{
% Overview of the MedMMV framework as a unified trajectory-level control system.
% \emph{Left (Inference-time)}: Given a patient case, MedMMV generates diverse diagnostic trajectories via multi-round short rollouts (Stage~1), repairs them in parallel using a multimodal evidence graph under hallucination detection (Stage~2), and selects the final diagnosis with uncertainty-aware CU scoring (Stage~3).\emph{Right (Training-time amortization)}: The same three control primitives define the post-training pipeline: Stage~1 provides candidate rollouts, Stage~2's repair signals annotate trajectory quality, and Stage~3's CU scores serve as a reward for on-policy distillation, enabling efficient student models to internalize trajectory preferences without online search.\TODO{draft drawn by Gemini, comment if have suggestions}
Overview of MedMMV as a trajectory-level control and amortized selection framework.
\emph{Left}: MedMMV performs inference-time trajectory control via proposal, evidence-grounded repair, and CU-based selection. 
\emph{Right}: The same pipeline produces trajectory-level signals that are converted into supervision through reject sampling, preference learning, and on-policy reranking. By optimizing likelihood over preferred trajectories, the student model amortizes trajectory selection into single-pass generation.
}
\label{fig:pipeline}
\vspace{-4mm}
\end{figure*}

\subsection{Stage 1: Trajectory Proposal}
\label{sec:stage1}
We generate $k$ independent short rollouts using a base MLLM, producing a candidate set ${p^{(1)}{\text{init}}, \dots, p^{(k)}{\text{init}}}$. This stage constructs the \emph{decision space} by approximating a distribution over plausible reasoning trajectories, rather than committing to a single path.

\noindent\textbf{Signal.} The candidate set exposes diverse reasoning alternatives and defines the space over which trajectory selection operates.

\subsection{Stage 2: Evidence-Grounded Trajectory Repair}
\label{sec:stage2}
Each trajectory is refined using an evidence graph $\mathcal{E}$ built from $\mathcal{T}$ and $\mathcal{I}$ via \textbf{TextDoctor} and \textbf{ImageDoctor}, and augmented with external medical knowledge retrieved by \textbf{WebSearch}. A hallucination and consistency detector iteratively verifies each step, triggering targeted revisions when inconsistencies are detected. This produces a refined trajectory $p_{\text{final}}$ and its supporting subgraph $\mathcal{E}_p$.

\noindent\textbf{Signal.} This stage provides step-level feedback on trajectory quality, identifying unsupported reasoning steps and capturing how trajectories evolve under correction.

\subsection{Stage 3: Uncertainty-Aware Trajectory Selection}
\label{sec:stage3}
We select the most reliable trajectory using a Combined Uncertainty (CU) score that aggregates evidence alignment, logical coherence, and repair cost. The CU score induces a ranking over candidate trajectories, explicitly modeling their relative reliability.

\noindent\textbf{Signal.} The resulting ranking provides trajectory-level preference supervision, directly encoding which reasoning paths should be selected over alternatives.

\subsection{Learning Trajectory Selection from Signals}
\label{sec:method_amortization}

While MedMMV improves reasoning through explicit multi-path control, its primary contribution lies in the supervision signals it produces over trajectory quality. These signals encode which reasoning paths are reliable, how they are corrected, and how they are ultimately selected.
We transfer these signals to a smaller model through post-training by converting them into training objectives that approximate the trajectory selection process. The key idea is to \emph{amortize} multi-path selection into a single-pass policy: instead of explicitly generating and comparing multiple trajectories, the student model learns to assign higher likelihood to trajectories that would be preferred by the MedMMV pipeline. Concretely, we derive supervision from the three stages of MedMMV:

\noindent\textbf{(1) Trace-level supervision (proposal + repair).}
We train the student to imitate teacher trajectories $p_{\text{final}}$, which already incorporate evidence-grounded corrections. This transfers local reasoning patterns and exposes the model to high-quality trajectories, but does not explicitly encode which trajectories are better than alternatives.

\noindent\textbf{(2) Selection-level supervision (trajectory ranking).}
The CU-based ranking induces preferences over candidate trajectories. We convert this signal into two forms of supervision, where trajectories are first sampled by an initial student model and then reranked in an offline manner: 
(i) \emph{rejection sampling}, where only top-ranked trajectories are retained for training, implicitly biasing the model toward preferred reasoning paths; 
(ii) \emph{pairwise preference learning}, where chosen and rejected trajectories define a ranking objective that directly optimizes the model to prefer higher-quality trajectories. These objectives explicitly approximate the selection stage by teaching the model to discriminate between competing reasoning paths.

\noindent\textbf{(3) On-policy supervision.}
We further expose the student to its own generated trajectories, which are evaluated and re-ranked by the teacher pipeline. The resulting top-ranked trajectories are used as training targets, aligning supervision with the student’s generation distribution and reinforcing selection behavior on realistic model outputs.

Through these mechanisms, the student learns to approximate the trajectory selection function implicitly: at inference time, it produces a single trajectory that reflects both the structure of high-quality reasoning and the preferences induced by the selection signals, without requiring explicit multi-path search.
Detailed training and implementation settings are provided in Appendix~\ref{app:posttrain_settings}.

\section{Experiments}

We evaluate MedMMV from a unified perspective: whether trajectory-level supervision improves reasoning by enabling better selection among competing reasoning paths, and whether this capability can be transferred through post-training.

\subsection{Trajectory Control Improves Selection Reliability}
\label{sec:inference_time_results}

We first evaluate whether MedMMV improves reasoning at inference time, where all modules are activated online (including trajectory generation, repair, and selection), by enabling better trajectory selection rather than relying on model scaling or prompting strategies.

\noindent\textbf{Datasets.}
We adopt six benchmarks covering diverse modalities and task formats. For multimodal reasoning: \textbf{MedXpertQA-MM} ~\citep{zuo2025medxpertqabenchmarkingexpertlevelmedical}, \textbf{MedFrameQA} ~\citep{yu2025medframeqamultiimagemedicalvqa}, and \textbf{PathVQA} ~\citep{he2020pathvqa30000questionsmedical}. For text-only: \textbf{MedXpertQA-Text} ~\citep{zuo2025medxpertqabenchmarkingexpertlevelmedical}, \textbf{MedMCQA} ~\citep{pal2022medmcqalargescalemultisubject}, and \textbf{MedQA} ~\citep{jin2020diseasedoespatienthave}. \footnote{Due to computational cost, we evaluate on stratified representative subsets that preserve the diversity of question types, difficulty levels, and body systems (Appendix~\ref{sec:dataset_stat_appendix}).}

\noindent\textbf{Metrics.}
\emph{Accuracy} as the proportion of correct final choices.
\emph{TRUE / INFO (T, I)} denote proxy truthfulness and informativeness scores, averaged over a three-model LLM judge ensemble ($\mathcal M={\text{GPT-4o}, \text{Claude-Sonnet-4}, \text{Gemini}}$)~\citep{lin2022truthfulqameasuringmodelsmimic, wang2025act}, with T$\times$I as a joint quality metric\footnote{T: Truthfulness (1–5); I: Informativeness (1–5); T×I: normalized product in $[0, 100]$, computed as $(T \times I)/25 \times 100$.}. While human annotation remains the gold standard, we use the LLM ensemble for scalability and complement it with targeted physician validation.

\noindent\textbf{Setup.}
We compare against: (1) \textit{Direct prompting models}, including GPT-5~\citep{gpt_5}, GPT-4o~\citep{gpt_4o}, GPT-oss-120B~\citep{gpt_oss}, Claude-Sonnet-4~\citep{anthropic_claude4_sonnet}, Qwen2.5-VL-7B, and Qwen2.5-VL-72B~\citep{bai2025qwen25vltechnicalreport}, prompted with zero-shot Chain-of-Thought (CoT); (2) \textit{Agent-based systems}, including MDAgents~\citep{kim2024mdagentsadaptivecollaborationllms}, ReConcile~\citep{chen2024reconcileroundtableconferenceimproves}, ColaCare~\citep{wang2025colacare}, and MedAgents~\citep{tang-etal-2024-medagents}.
For modular comparisons, each MedMMV variant uses the same MLLM backbone as its paired CoT baseline. Each trajectory allows up to three self-revision loops, and all agent baselines use GPT-5 as the backbone for fair comparison.

\begin{table*}[!ht]
\vspace{-2mm}
\centering
\tiny
\begin{tabular}{@{}llcccccc@{}}
\toprule
\multirow{2}{*}{\textbf{Category}} & \multirow{2}{*}{\textbf{Methods}} & \multicolumn{3}{c}{\textbf{Medical VQA}} & \multicolumn{3}{c}{\textbf{Medical QA}} \\
\cmidrule(lr){3-5} \cmidrule(lr){6-8}
& & \makecell{MedXpert-MM \\ (\TextCircle\ImageCircle)} & \makecell{MedFrameQA \\ (\TextCircle\ImageCircle)} & \makecell{PathVQA \\ (\TextCircle\ImageCircle)} & \makecell{MedXpert-Text \\ (\TextCircle)} & \makecell{MedQA \\ (\TextCircle)} & \makecell{MedMCQA \\ (\TextCircle)} \\
\midrule
\multirow{6}{*}{CoT Baselines} 
& GPT-5             & \cellcolor{red!5}{\strut 67.90\stddev{2.19}} & 46.95\stddev{2.87} & \cellcolor{red!5}{\strut 67.91\stddev{3.10}} & \cellcolor{red!5}{\strut 32.40\stddev{2.12}} & \cellcolor{red!20}{\strut 99.07\stddev{0.72}} & 35.34\stddev{3.50} \\
& GPT-4o            & 37.28\stddev{1.98} & 42.62\stddev{3.13} & 37.03\stddev{3.08} & 19.49\stddev{1.84} & 94.81\stddev{1.47} & 32.39\stddev{3.42} \\
& GPT-oss-120B      & 50.03\stddev{1.98} & 45.90\stddev{2.92} & 44.81\stddev{3.61} & 20.66\stddev{1.87} & 96.96\stddev{1.19} & 33.36\stddev{3.35} \\
& Claude-Sonnet-4   & 46.82\stddev{2.08} & 47.64\stddev{3.45} & 65.81\stddev{3.67} & 23.31\stddev{2.27} & 93.82\stddev{1.64} & 30.33\stddev{3.43} \\
& Qwen2.5-VL-7B     & 10.50\stddev{5.48} & 20.31\stddev{7.01} & 49.21\stddev{4.33} & 10.67\stddev{8.00} & 48.99\stddev{3.47} & 26.52\stddev{3.00} \\
& Qwen2.5-VL-72B    & 35.96\stddev{2.34} & 42.33\stddev{3.18} & 66.01\stddev{3.36} & 12.57\stddev{1.52} & 74.06\stddev{3.33} & 27.28\stddev{3.10} \\
\hdashline
\multirow{4}{*}{Medical Agents} 
& MDAgents          & 44.21\stddev{3.57} & 39.36\stddev{3.45} & 64.66\stddev{3.25} & 17.48\stddev{2.58} & 81.08\stddev{2.62} & 31.02\stddev{3.76} \\
& ReConcile         & 47.31\stddev{3.16} & 45.12\stddev{4.05} & 73.31\stddev{3.27} & 17.91\stddev{2.36} & 89.85\stddev{2.27} & 29.31\stddev{3.13} \\
& ColaCare          & 33.20\stddev{3.45} & 39.06\stddev{3.42} & 72.19\stddev{3.15} & 14.96\stddev{2.31} & 88.93\stddev{2.63} & 29.05\stddev{3.27} \\
& MedAgent          & 36.87\stddev{3.42} & 44.17\stddev{3.40} & 70.20\stddev{2.98} & 17.76\stddev{2.46} & 89.70\stddev{2.10} & 31.09\stddev{3.14} \\
\hdashline
\multirow{6}{*}{MedMMV}
& GPT-5             & \cellcolor{red!35}{\strut 73.58\stddev{1.87}} & \cellcolor{red!20}{\strut 59.67\stddev{1.76}} & \cellcolor{red!35}{\strut 68.25\stddev{2.08}} & \cellcolor{red!35}{\strut 39.46\stddev{1.08}} & \cellcolor{red!35}{\strut 99.15\stddev{3.00}} & \cellcolor{red!20}{\strut 39.62\stddev{2.69}} \\
& GPT-4o            & 60.24\stddev{2.33} & 53.08\stddev{2.45} & 39.46\stddev{3.53} & 26.31\stddev{3.31} & 94.62\stddev{3.53} & 38.64\stddev{3.13} \\
& GPT-oss-120B      & \cellcolor{red!20}{\strut 69.18\stddev{0.91}} & \cellcolor{red!35}{\strut 67.64\stddev{1.89}} & 50.15\stddev{3.88} & 28.62\stddev{3.98} & \cellcolor{red!5}{\strut 97.15\stddev{4.01}} & \cellcolor{red!35}{\strut 42.46\stddev{3.18}} \\
& Claude-Sonnet-4   & 61.65\stddev{2.77} & \cellcolor{red!5}{\strut 57.92\stddev{3.69}} & \cellcolor{red!20}{\strut 68.08\stddev{3.76}} & \cellcolor{red!20}{\strut 32.77\stddev{3.57}} & 96.42\stddev{3.53} & 32.31\stddev{3.07} \\
& Qwen2.5-VL-7B     & 35.35\stddev{2.34} & 40.15\stddev{6.01} & 51.15\stddev{4.29} & 18.62\stddev{6.54} & 56.08\stddev{4.78} & \cellcolor{red!5}{\strut 36.03\stddev{6.23}} \\
& Qwen2.5-VL-72B    & 45.62\stddev{2.32} & 47.40\stddev{4.67} & 66.62\stddev{3.85} & 20.46\stddev{2.92} & 85.92\stddev{5.86} & 31.46\stddev{4.70} \\
\bottomrule
\end{tabular}
% \captionsetup{justification=raggedright,singlelinecheck=false,font=tiny}
\vspace{-2.5mm}
\caption{Overall accuracy results on six medical benchmarks. \TextCircle: text modality; \ImageCircle: image modality. Red shading highlights the top three methods for each dataset. 
% Accuracy is reported as mean (\%) with standard deviation across multiple runs.
}
\label{tab:main_results}
\vspace{-4mm}
\end{table*}
\begin{table*}[!ht]
\centering
\tiny
\setlength{\tabcolsep}{2pt}
\begin{tabular}{@{}ll*{18}{c}@{}}
\toprule
\multirow{3}{*}{\textbf{Category}} & \multirow{3}{*}{\textbf{Methods}} & \multicolumn{9}{c}{\textbf{Medical VQA}} & \multicolumn{9}{c}{\textbf{Medical QA}} \\
\cmidrule(lr){3-11} \cmidrule(lr){12-20}
& & \multicolumn{3}{c}{\makecell{MedXpert-MM \\ (\TextCircle\ImageCircle)}} & \multicolumn{3}{c}{\makecell{MedFrameQA \\ (\TextCircle\ImageCircle)}} & \multicolumn{3}{c}{\makecell{PathVQA \\ (\TextCircle\ImageCircle)}} & \multicolumn{3}{c}{\makecell{MedXpert-Text \\ (\TextCircle)}} & \multicolumn{3}{c}{\makecell{MedQA \\ (\TextCircle)}} & \multicolumn{3}{c}{\makecell{MedMCQA \\ (\TextCircle)}} \\
\cmidrule(lr){3-5} \cmidrule(lr){6-8} \cmidrule(lr){9-11} \cmidrule(lr){12-14} \cmidrule(lr){15-17} \cmidrule(lr){18-20}
& & T & I & T×I & T & I & T×I & T & I & T×I & T & I & T×I & T & I & T×I & T & I & T×I \\
\midrule
\multirow{6}{*}{CoT Baselines} 
& GPT-5           & \cellcolor{red!5}{\strut 4.17} & \cellcolor{red!20}{\strut 4.78} & 72.03 & 3.44 & \cellcolor{red!5}{\strut 4.57} & 65.84 & 3.51 & \cellcolor{red!5}{\strut 4.59} & 68.00 & \cellcolor{red!20}{\strut 3.68} & \cellcolor{red!20}{\strut 4.70} & \cellcolor{red!20}{\strut 71.36} & \cellcolor{red!5}{\strut 4.92} & \cellcolor{red!35}{\strut 4.98} & \cellcolor{red!20}{\strut 98.30} & 2.67 & \cellcolor{red!5}{\strut 4.56} & \cellcolor{red!20}{\strut 51.00} \\
& GPT-4o          & 3.31 & 4.57 & 56.47 & 3.23 & 4.24 & 57.40 & 3.51 & 3.54 & 58.67 & 2.82 & 4.32 & 51.98 & 4.72 & 4.90 & 93.82 & 2.56 & 4.51 & 48.53 \\
& GPT-oss-120B    & 3.58 & 4.65 & 58.82 & 3.54 & \cellcolor{red!20}{\strut 4.61} & \cellcolor{red!5}{\strut 68.10} & 3.10 & 3.82 & 52.34 & 2.97 & 4.30 & 55.57 & 4.87 & \cellcolor{red!20}{\strut 4.97} & 97.18 & 2.55 & 4.44 & 47.69 \\
& Claude-Sonnet-4           & 3.76 & 4.69 & 63.37 & 3.11 & 4.42 & 58.17 & 3.18 & 4.54 & 61.26 & 2.99 & 4.46 & 56.23 & 4.64 & 4.90 & 92.15 & 2.43 & 4.48 & 45.57 \\
& Qwen2.5-VL-7B   & 1.68 & 2.53 & 18.22 & 2.12 & 2.23 & 24.44 & 2.74 & 3.08 & 39.14 & 1.20 & 2.10 & 11.15 & 2.81 & 2.65 & 36.12 & 1.79 & 2.50 & 21.03 \\
& Qwen2.5-VL-72B  & 2.61 & 4.18 & 46.86 & 2.81 & 4.26 & 51.27 & 3.44 & \cellcolor{red!35}{\strut 4.62} & 63.28 & 2.18 & 3.76 & 34.95 & 3.78 & 4.19 & 69.26 & 2.36 & 3.80 & 38.99 \\
\hdashline
\multirow{4}{*}{Medical Agents}
& MDAgents        & 2.11 & 1.84 & 17.19 & 3.08 & 4.50 & 58.32 & 3.30 & \cellcolor{red!20}{\strut 4.60} & 65.19 & 2.08 & 3.78 & 44.32 & 4.33 & 4.71 & 85.41 & 2.31 & 4.00 & 45.23  \\
& ReConcile       & 3.13 & 4.21 & 56.63 & 3.43 & 4.48 & 65.11 & \cellcolor{red!20}{\strut 3.99} & 4.12 & \cellcolor{red!20}{\strut 70.88} & 2.17 & 3.68 & 32.76 & 4.60 & 4.86 & 90.38 & \cellcolor{red!5}{\strut 2.73} & 3.86 & 42.51 \\
& ColaCare        & 2.72 & 4.09 & 48.44 & 3.39 & 4.52 & 52.27 & 3.49 & 4.19 & \cellcolor{red!35}{\strut 72.10} & 2.39 & 3.63 & 35.27 & 4.64 & 4.79 & 90.16 & 2.61 & 3.70 & 37.82 \\
& MedAgent        & 3.41 & 4.62 & 52.89 & 3.37 & 4.55 & 60.50 & 3.41 & 4.33 & \cellcolor{red!5}{\strut 70.63} & 2.58 & 3.81 & 38.22 & 4.59 & 4.77 & 88.61 & 2.56 & 3.84 & 46.03 \\
\hdashline
\multirow{6}{*}{MedMMV}
& GPT-5           & \cellcolor{red!35}{\strut 4.26} & \cellcolor{red!5}{\strut 4.73} & \cellcolor{red!35}{\strut 82.89} & \cellcolor{red!20}{\strut 4.00} & 4.42 & \cellcolor{red!20}{\strut 74.22} &  3.74 & 4.02 & 65.36 & \cellcolor{red!35}{\strut 4.22} & \cellcolor{red!20}{\strut 4.70} & \cellcolor{red!35}{\strut 81.53} & \cellcolor{red!20}{\strut 4.94} & 4.82 & 95.36 & \cellcolor{red!35}{\strut 2.97} & \cellcolor{red!20}{\strut 4.57} & \cellcolor{red!35}{\strut 56.40} \\
& GPT-4o          & \cellcolor{red!20}{\strut 4.20} & 4.42 & \cellcolor{red!5}{\strut 76.70} & \cellcolor{red!5}{\strut 3.72} & 3.90 & 60.00 & 3.61 & 3.76 & 59.58 & 3.55 & 4.05 & 60.22 & \cellcolor{red!35}{\strut 4.95} &  \cellcolor{red!5}{\strut 4.95} & \cellcolor{red!5}{\strut 97.96} & 2.58 & \cellcolor{red!35}{\strut 4.93} & 49.02 \\
& GPT-oss-120B    &  4.01 & \cellcolor{red!35}{\strut 4.84} & \cellcolor{red!20}{\strut 80.95} & \cellcolor{red!35}{\strut 4.27} & \cellcolor{red!35}{\strut 4.77} & \cellcolor{red!35}{\strut 74.71} & \cellcolor{red!35}{\strut 4.03} & 3.57 & 61.96 & \cellcolor{red!5}{\strut 3.63} & \cellcolor{red!35}{\strut 4.77} & \cellcolor{red!5}{\strut 69.71} & \cellcolor{red!35}{\strut 4.95} & \cellcolor{red!35}{\strut 4.98} & \cellcolor{red!35}{\strut 98.73} & \cellcolor{red!20}{\strut 2.94} & 4.31 & \cellcolor{red!5}{\strut 50.94} \\
& Claude-Sonnet-4 & 3.81 & 4.16 & 67.41 & 3.54 & 3.98 & 57.87 & \cellcolor{red!5}{\strut 3.78} & 3.18 & 68.55 & 3.65 & \cellcolor{red!5}{\strut 4.57} & 69.36 & 4.91 & \cellcolor{red!5}{\strut 4.95} & 97.54 & 2.47 & 4.25 & 49.82 \\
& Qwen2.5-VL-7B   & 2.59 & 2.59 & 30.06 & 3.33 & 3.30 & 45.76 & 3.27 & 3.52 & 46.02 & 2.28 & 3.08 & 20.42 & 3.70 & 3.72 & 47.85 & 2.13 & 2.72 & 23.71 \\
& Qwen2.5-VL-72B  & 3.12 & 4.35 & 54.80 & 3.42 & 4.22 & 58.80 & 3.72 & 4.40 & 69.47 & 2.60 & 3.10 & 39.29 & 4.32 & 4.03 & 76.71 & 2.55 & 3.91 & 41.02 \\
\bottomrule
\end{tabular}
% \captionsetup{justification=raggedright,singlelinecheck=false,font=tiny}
\vspace{-2mm}
\caption{Truthfulness (T), informativeness (I), and joint quality (T×I) on benchmarks. \TextCircle: text modality; \ImageCircle: image modality. Red shading marks the top three methods per dataset. 
% Scores are averaged over three independent LLM judges (GPT-5, Claude-Sonnet-4, Gemini), which helps mitigate single-model bias and prevents inflation from models rating their own outputs.
}
\label{tab:tri_metrics_results}

\vspace{-3mm}
\end{table*}

\noindent\textbf{Main results.}
Table~\ref{tab:main_results} shows consistent improvements across all benchmarks. With GPT-5, MedMMV improves accuracy from 67.9\% to 73.6\% on MedXpert-MM and from 32.4\% to 39.5\% on MedXpert-Text, with larger gains on complex multimodal tasks such as MedFrameQA (+12.7\%). Because the backbone is unchanged, these improvements isolate trajectory-level control as the primary driver. This supports our hypothesis that performance gains arise from better selection among competing reasoning trajectories.

\noindent\textbf{Reliability analysis.}
Table~\ref{tab:tri_metrics_results} shows that MedMMV improves truthfulness without sacrificing informativeness. For example, TRUE increases from 4.17 to 4.26 (MedXpert-MM) and from 3.68 to 4.22 (MedXpert-Text), yielding T$\times$I gains of +10.9\% and +10.2\%. Notably, baseline MLLMs and medical agents often achieve high INFO yet lag in TRUE; across datasets, INFO remains stable ($\approx$ 4.5–4.7), indicating that gains come from more accurate, not merely more verbose reasoning. These results indicate that trajectory-level control improves reliability by favoring better-supported reasoning paths.

\begin{table}[!t]
\vspace{-2mm}
\centering
\scriptsize
\setlength{\tabcolsep}{4pt}
\resizebox{\columnwidth}{!}{%
\begin{tabular}{@{}l c c c c c c@{}}
\toprule
\multirow{2}{*}{\textbf{Body System }} & \multicolumn{3}{c}{\textbf{MedMMV}} & \multicolumn{3}{c}{\textbf{CoT}} \\
\cmidrule(lr){2-4} \cmidrule(l){5-7}
& \textbf{T} & \textbf{I} & \textbf{T$\times$I} & \textbf{T} & \textbf{I} & \textbf{T$\times$I} \\
\midrule
Skeletal & 4.42\stddev{0.34} & 4.13\stddev{0.40} & 73.26\stddev{12.87} & 4.11\stddev{0.17} & 4.47\stddev{0.38} & 73.39\stddev{6.12} \\
Reproductive & 4.75\stddev{0.23} & 4.00\stddev{0.01} & 76.17\stddev{3.60} & 3.60\stddev{0.40} & 4.66\stddev{0.27} & 67.23\stddev{11.38} \\
Cardiovascular & 4.28\stddev{0.46} & 3.97\stddev{0.26} & 68.28\stddev{11.95} & 3.15\stddev{0.58} & 4.02\stddev{0.33} & 51.12\stddev{13.13} \\
Lymphatic & 4.23\stddev{0.45} & 4.18\stddev{0.39} & 71.28\stddev{13.83} & 3.87\stddev{0.60} & 3.98\stddev{0.67} & 62.69\stddev{20.83} \\
Nervous & 4.27\stddev{0.55} & 3.68\stddev{0.30} & 63.31\stddev{13.41} & 3.18\stddev{0.42} & 4.23\stddev{0.63} & 54.59\stddev{15.35} \\
Digestive & 4.19\stddev{0.64} & 3.69\stddev{0.30} & 62.46\stddev{14.07} & 3.32\stddev{0.20} & 4.03\stddev{0.44} & 53.56\stddev{7.16} \\
Urinary & 4.45\stddev{0.43} & 4.03\stddev{0.33} & 72.05\stddev{11.95} & 3.73\stddev{0.70} & 4.07\stddev{0.35} & 61.37\stddev{16.28} \\
Endocrine & 4.22\stddev{0.29} & 3.63\stddev{0.12} & 61.31\stddev{5.22} & 3.02\stddev{0.70} & 3.95\stddev{0.22} & 47.35\stddev{9.07} \\
Integumentary & 4.43\stddev{0.42} & 4.11\stddev{0.29} & 73.07\stddev{11.04} & 3.44\stddev{0.09} & 4.14\stddev{0.40} & 57.03\stddev{5.82} \\
\midrule
\textbf{Average} & \textbf{4.36\stddev{0.17}} & \textbf{3.94\stddev{0.20}} & \textbf{69.02\stddev{5.12}} & \textbf{3.49\stddev{0.34}} & \textbf{4.17\stddev{0.23}} & \textbf{58.70\stddev{7.77}} \\
\bottomrule
\end{tabular}%
}
\vspace{-2mm}
\caption{Physician evaluation on the MedXpert-MM dataset by body system. Each body system includes 20 samples; the per-doctor breakdown is in Appendix~\ref{sec:all_human_eval_table_appendix}.}
\label{tab:medxpert_mm_by_system}
\vspace{-6mm}
\end{table}
\begin{table*}[!t]
% \vspace{-2mm}
\centering
\scriptsize
\setlength{\tabcolsep}{2pt}
\begin{tabular}{@{}ll*{8}{c}@{}}
\toprule
\multirow{3}{*}{\textbf{Category}} & \multirow{3}{*}{\textbf{Methods}} & \multicolumn{4}{c}{\textbf{MedXpert-MM}} & \multicolumn{4}{c}{\textbf{MedXpert-Text}} \\
\cmidrule(lr){3-6} \cmidrule(lr){7-10}
& & \multicolumn{4}{c}{\makecell{(\TextCircle\ImageCircle)}} & \multicolumn{4}{c}{\makecell{(\TextCircle)}} \\
\cmidrule(lr){3-6} \cmidrule(lr){7-10}
& & Acc & T & I & T×I & Acc & T & I & T×I \\
\midrule
\multirow{1}{*}{Full model} 
& \textbf{MedMMV (GPT-5)} & \textbf{73.58} & \textbf{4.26} & \textbf{4.73} & \textbf{82.89} & \textbf{39.46} & \textbf{4.22} & \textbf{4.70} & \textbf{81.53} \\
\hdashline
\multirow{2}{*}{Core framework} 
&  w/o Path Expansion with multi-round short CoT & 68.45 & 4.23 & 4.70 & 79.52 & 36.15 & 4.15 & 4.65 & 77.80 \\
&  w/o Self-feedback halluc. detector & 69.80 & 4.19 & 4.68 & 76.45 & 35.50 & 4.02 & 4.62 & 74.20 \\
\hdashline
\multirow{3}{*}{Expert agents} 
&  w/o TextDoctor agent & 72.10 & 4.24 & 4.71 & 80.15 & 38.25 & 4.18 & 4.68 & 79.10 \\
&  w/o ImageDoctor agent & 72.95 & 4.25 & 4.72 & 81.60 & - & - & - & - \\
&  w/o WebSearch agent & 73.15 & 4.25 & 4.72 & 82.05 & 38.90 & 4.20 & 4.69 & 80.45 \\
\hdashline
\multirow{1}{*}{Decision strategy} 
&  Random decision instead of CU scoring & 68.55 & 4.18 & 4.65 & 73.90 & 33.25 & 3.85 & 4.58 & 71.85 \\
\bottomrule
\end{tabular}
% \captionsetup{justification=raggedright,singlelinecheck=false,font=tiny}
\vspace{-2mm}
\caption{Ablation study on MedXpert-MM and MedXpert-Text. Full ablations on six benchmarks are in Table~\ref{tab:full_ablation_medqa}.
% \textit{\textbf{Note:}} Acc: Accuracy (\%); T: Truthfulness (1–5); I: Informativeness (1–5); T×I: normalized product in [0,100], computed as $(T \times I)/25 \times 100$. 
% ImageDoctor ablations are not applicable (–) for text-only tasks. Bold indicates full MedMMV with best performance.
}
\label{tab:ablation_medqa}
\vspace{-4mm}
\end{table*}

\noindent\textbf{Physician evaluation.}
To complement automated metrics, we conducted a rigorous human evaluation on MedXpert-MM with 27 physicians to assess the clinical quality of generated responses.
As shown in Table~\ref{tab:medxpert_mm_by_system}, MedMMV is consistently preferred over CoT (both using GPT-5), improving Truthfulness from 3.49 to 4.36 and T$\times$I from 58.70 to 69.02. Importantly, these human judgments closely align with our automatic LLM-based evaluation, indicating that the LLM-as-judge framework provides a reliable proxy for expert assessment. This consistency supports the use of automatic judges for scalable evaluation while preserving alignment with domain expert preferences.

\noindent\textbf{Ablation study.}
Table~\ref{tab:ablation_medqa} summarizes the key ablations. Replacing the CU-based selector with random selection causes the largest degradation, reducing accuracy by $\sim$5\% on MedXpert-MM ($\sim$6\% on the text-only split) and TRUE$\times$INFO by $\sim$9\%, clearly identifying trajectory selection (Stage~3) as the dominant contributor. In comparison, removing the hallucination detector in Stage~2 lowers accuracy by $\sim$4\%, and disabling path expansion leads to further drops of $\sim$5\% on multimodal tasks and $\sim$3\% on text-only tasks. Removing specialist agents results in only modest changes (1--4\%). These results collectively show that while repair and auxiliary modules provide incremental gains, the primary performance improvements stem from selecting among competing trajectories, directly supporting our central claim that trajectory selection is the key driver of reliable reasoning\footnote{Additional case studies are provided in Appendix~\ref{sec:detailed_results_appendix} and Figure~\ref{fig:case_study_fig_appendix}.}.

\subsection{Learning Trajectory Selection from Signals}
\label{sec:post_training}
The above results suggest that MedMMV’s gains stem from improved trajectory selection. We next ask if selection capability can be transferred to a smaller model through supervision.

\noindent\textbf{Setup.}
We train Qwen2.5-VL-7B as the student using MedMMV (GPT-5) as the teacher on MedXpertQA-MM. We compare a sequence of post-training variants that transfer progressively richer pipeline signals to the student. Specifically, we consider three classes of supervision. 
(1) \textbf{Trace imitation}, where the student learns to mimic reasoning trajectories, including \textit{Trace-Imitation (Raw-CoT)} and \textit{Trace-Imitation (Teacher-Trajectory)} using MedMMV-generated traces. 
(2) \textbf{Offline trajectory selection}, where trajectory quality is enforced via rejection sampling or preference learning. This includes \textit{RS-Imitation (Self-Consistency)} and \textit{RS-Imitation (CU-Select)}, which perform rejection sampling based on different selection signals, as well as \textit{Pairwise-Preference (Offline)}, which introduces explicit chosen/rejected supervision. RS-Imitation can be viewed as implicit preference learning, where high-quality trajectories are selected via rejection sampling and used for supervised training.
(3) \textbf{On-policy feedback}, where the student learns from its own rollouts, including \textit{OnPolicy-RS (Self-Rollout)} and \textit{OnPolicy-Preference (Self-Rollout + Repair)}.

Shown in Table~\ref{tab:signal_ladder}, the compared variants differ along four axes: whether training uses filtered trajectories, pairwise preference supervision, student-on-distribution rollouts, and repair-aware scoring. All on-policy variants are initialized from SFT-trained models for stable optimization. We use 134 teacher-annotated samples for training and evaluate on a disjoint held-out split. All variants share the matched training budgets and identical LoRA configurations for fair comparison. Full details are deferred to Appendix~\ref{app:posttrain_settings}.

\begin{table*}[!t]
\centering
\scriptsize
\setlength{\tabcolsep}{4pt}
\renewcommand{\arraystretch}{1.12}
\vspace{-2mm}
\begin{tabular}{@{}lcccccccc@{}}
\toprule
\textbf{Method} & \textbf{Filtered} & \textbf{Pairwise} & \textbf{On-dist.} & \textbf{Repair} & \textbf{Acc} & \textbf{T} & \textbf{I} & \textbf{T$\times$I} \\
\midrule

\multicolumn{9}{@{}l}{\textbf{Imitation}} \\
Base (no training)      & \xmark & \xmark   & \xmark & \xmark   & 10.50 & 1.68 & 2.53 & 18.22 \\
Trace-Imitation (raw CoT)     & \xmark & \xmark   & \xmark & \xmark   & 20.00 & 1.93  & 2.79 & 23.41 \\
Trace-Imitation (Teacher-Trajectory)      & \xmark & \xmark   & \xmark & partial  & 24.00 & 2.18 & 2.81 & 26.18 \\
\hdashline

\multicolumn{9}{@{}l}{\textbf{Preference}} \\
RS-Imitation (Self-Consistency)                  & weak   & \xmark   & \xmark & \xmark   & 23.00 & 2.08 & 2.90 & 26.00 \\
RS-Imitation (CU-Select)   & \cmark & \xmark   & \xmark & partial  & 23.00 & 2.16 & 2.74 & 25.68 \\
Pairwise-Preference (Offline)            & \cmark & \cmark   & \xmark & partial  & \textbf{29.00} & 2.34 & 2.89 & 28.09 \\
\hdashline

\multicolumn{9}{@{}l}{\textbf{On-policy}} \\
OnPolicy-RS (Self-Rollout)             & \cmark & implicit & \cmark & \xmark   & 24.00 & 2.13 & 2.84 & 27.61 \\
OnPolicy-Preference (Self-Rollout + Repair)           & \cmark & implicit & \cmark & \cmark   & 26.00 & 2.25 & 2.76 & 27.62 \\

\bottomrule
\end{tabular}
\vspace{-2mm}

\caption{Signal ladder: from trace imitation to closed-loop feedback transfer on MedXpert-MM (Qwen2.5-VL-7B student, MedMMV GPT-5 teacher). Each row adds a richer pipeline signal. \emph{Filtered}: CU-based trajectory filtering; \emph{Pairwise}: chosen/rejected preference objective; \emph{On-dist.}: student generates rollouts; \emph{Repair}: repair-aware scoring.}
\label{tab:signal_ladder}
\vspace{-4mm}
\end{table*}
\begin{figure*}[!t]
\centering
\includegraphics[width=\linewidth]{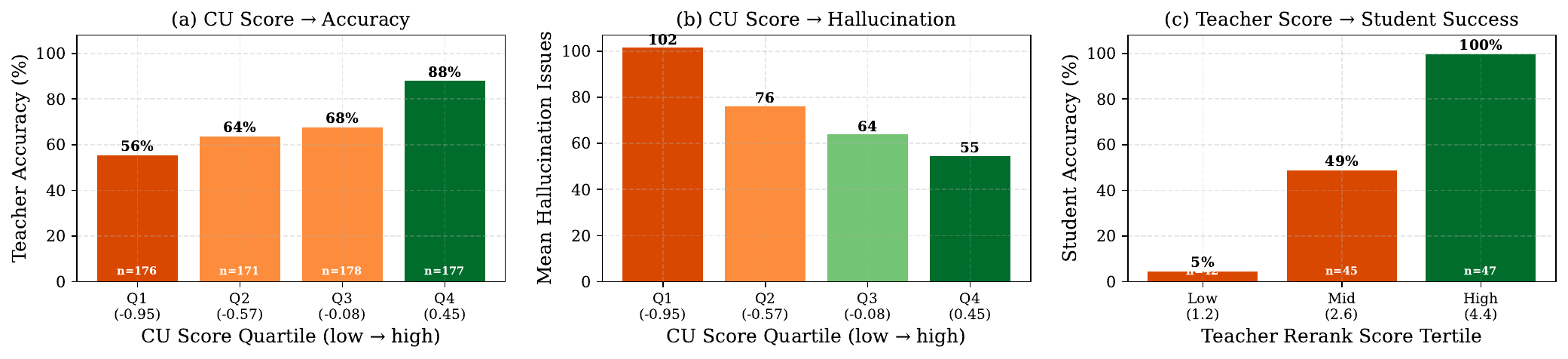}
\vspace{-6mm}
% \caption{
% Quality calibration of pipeline scores.
% \emph{(a)}~CU score is monotonically associated with teacher accuracy (56\%$\to$88\% from Q1 to Q4), confirming that CU reflects genuine reasoning quality.
% \emph{(b)}~Higher CU quartiles exhibit fewer hallucination issues (102$\to$55), validating CU as an evidence-grounded quality signal.
% \emph{(c)}~Teacher reranking scores on student rollouts are strongly predictive of student success (5\%$\to$100\% accuracy from low to high tertile), demonstrating that the teacher provides a calibrated, not arbitrary, reward signal for on-policy learning.
% }
\caption{
Pipeline feedback signals are calibrated for reasoning quality.
(a) CU score increases with teacher accuracy, indicating alignment with true trajectory quality.
(b) Higher-CU trajectories exhibit fewer hallucination issues, demonstrating that CU captures factual grounding.
(c) Teacher reranking scores strongly predict student success, establishing teacher feedback as an effective reward signal for on-policy learning.
}
\label{fig:signal_calibration}
\vspace{-6mm}
\end{figure*}

\noindent\textbf{Main results: selection signals transfer.}
Table~\ref{tab:signal_ladder} shows a clear progression: as supervision moves from trace imitation to trajectory selection and preference learning, performance improves consistently. Starting from the base model, trace imitation improves performance to 20.00\% (raw CoT) and further to 24.00\% when using MedMMV-generated traces. Introducing preference-based supervision yields additional gains: Pairwise-Preference (Offline) peaks at 29.00\% accuracy, outperforming all imitation-based methods. On-policy methods further improve over weaker baselines (e.g., OnPolicy-Preference (Self-Rollout + Repair) reaches 26.00\%), suggesting that aligning supervision with the student’s own distribution can provide additional benefits in some cases. These results validate our central hypothesis that trajectory selection is a transferable capability. This is further supported by the calibration analysis in Figure~\ref{fig:signal_calibration}, where pipeline-derived signals exhibit strong alignment with trajectory correctness and hallucination, and are predictive of downstream student success.

\noindent\textbf{Offline vs. on-policy.}
We observe that on-policy methods do not provide clear advantages over offline training in this setting. For example, OnPolicy-Preference (Self-Rollout + Repair) achieves 26.00\% accuracy, which is slightly below the best offline variant (Pairwise-Preference (Offline), 29.00\%), and comparable to other offline methods. This is unlikely to be due to poor signal quality, as Appendix~\ref{app:trajectory_reward} shows that both CU scores and teacher reranking signals are well calibrated and strongly correlated with trajectory correctness.
One possible explanation is that the 7B model has limited capacity to fully utilize richer supervision signals introduced by on-policy training. While on-policy methods provide distribution-aligned and trajectory-level feedback, the resulting gains may be constrained if the model cannot effectively internalize and exploit these signals beyond what is already captured by offline supervision.
However, this observation is specific to the current setting, and further investigation is needed to understand how model scale affects the relative benefits of offline and on-policy learning.

\subsection{Reliability, Stability, and Practicality}
\label{sec:reliability_robustness}

We further evaluate robustness, reliability, and deployment properties (Table~\ref{tab:robustness_summary}; details in Appendix~\ref{app:reliability_details}). Physician ratings show strong agreement, with ICC(T)=0.89 and ICC(I)=0.85, and the physician-measured improvement closely matches that of the LLM judges ($\Delta$T$\times$I: 10.32 vs.\ 10.86), supporting the reliability of our LLM-as-judge evaluation. Under mixed perturbations to the evidence graph, MedMMV maintains accuracy above 72\% up to 40\% noise and then degrades smoothly, while varying CU weights changes accuracy by at most 1\%, indicating that neither the graph nor the scorer behaves as a brittle component. Despite explicit multi-path control, MedMMV remains tractable (40–62s per query with parallelization). For deployment-constrained settings, the learned model after post-training reaches most reasoning quality with standard single-pass generation, eliminating the need for multi-path control and evidence graph construction.

\section{Discussion and Conclusion}
We present MedMMV, an evidence-grounded trajectory control framework for reliable multimodal clinical reasoning. MedMMV constructs, repairs, and evaluates multiple reasoning trajectories, enabling robust decision-making under uncertainty. Beyond improving final accuracy, it produces structured supervision signals that capture trajectory reliability. Experiments show these signals strongly align with reasoning quality, with gains primarily driven by trajectory selection, which is confirmed by ablations and improved transfer under trajectory-level supervision. We further demonstrate that trajectory selection is a learnable capability: through post-training on candidate trajectories, repair feedback, and preference signals, smaller models learn to favor reliable reasoning paths without requiring multi-path control at inference. Notably, explicit preference supervision yields the largest improvements, highlighting that selecting trustworthy trajectories is more critical than imitating reasoning traces. Overall, our results suggest that improving reasoning hinges not only on generating trajectories but on selecting among them, establishing trajectory selection as a key transferable capability for reliable multimodal reasoning.

\section*{Limitations and Future Work.}
Despite the benefits of explicit trajectory control, several limitations remain. First, evaluation relies largely on model-based proxy metrics; although partially validated by physicians, these do not replace large-scale expert annotation or real-world clinical studies. Second, reasoning quality is bounded by the evidence graph: errors in extraction, retrieval, or construction can propagate, and the current framework lacks robust graph-level self-correction. Third, the effects of trajectory control and retrieval are not fully disentangled; while ablations suggest selection dominates, stronger retrieval-only and graph-only baselines are needed. Fourth, multi-path control incurs substantial inference cost, limiting applicability to non-time-critical settings. Finally, although trajectory selection transfers to smaller models via post-training, this has only been validated on a limited set of architectures and benchmarks.
Future work includes improving evidence-graph construction and self-repair, designing stronger baselines to isolate control from retrieval, reducing cost via adaptive trajectory expansion or early stopping, and expanding validation across datasets, model families, and clinically realistic scenarios.

\section*{Ethics Statement}
This work investigates the reliability of medical reasoning agents but is not intended for real-time primary clinical deployment. All datasets are public and de-identified, and no patient data was collected. Physician evaluations were voluntary and confined to blinded response ratings. While MedMMV reduces proxy hallucination metrics, errors may still occur, and safe application requires regulatory approval, human oversight, and further validation in real-world settings. We position MedMMV as a research prototype suitable for asynchronous second-opinion support, emphasizing that human oversight is essential for any clinical use.

% Acknowledgments and Ethics Statement are in Conclusion.tex

\bibliography{ref}
% \bibliography{custom}

\newpage

\appendix
% \section*{Appendix}

\section{Limitations and Future Work.}
Despite the benefits of explicit trajectory control, several limitations remain. First, evaluation relies largely on model-based proxy metrics; although partially validated by physicians, these do not replace large-scale expert annotation or real-world clinical studies. Second, reasoning quality is bounded by the evidence graph: errors in extraction, retrieval, or construction can propagate, and the current framework lacks robust graph-level self-correction. Third, the effects of trajectory control and retrieval are not fully disentangled; while ablations suggest selection dominates, stronger retrieval-only and graph-only baselines are needed. Fourth, multi-path control incurs substantial inference cost, limiting applicability to non-time-critical settings. Finally, although trajectory selection transfers to smaller models via post-training, this has only been validated on a limited set of architectures and benchmarks.
Future work includes improving evidence-graph construction and self-repair, designing stronger baselines to isolate control from retrieval, reducing cost via adaptive trajectory expansion or early stopping, and expanding validation across datasets, model families, and clinically realistic scenarios.

% \section{The Use of Large Language Models (LLMs)}
% The authors used a large language model (LLM) solely for language polishing and grammar correction. The LLM served strictly as a writing enhancement tool. The authors assume full responsibility for all content, ideas, and the final presentation of the work.

\section{Preliminary Analysis Details}
\label{app:analysis-details}

\subsection{Pilot Metrics and Implementation Details}
\label{app:pilot-metrics}

% \textbf{Base metrics.}
% For each question $q$ with $k_q$ generations, let $c_{q,i}\in\mathcal O_q$ be the option at round $i$.
% The empirical prior before round $i\ge2$ is
% \[
% \mathbf p_{q,i-1}(o)=\tfrac{1}{i-1}\sum_{t=1}^{i-1}\mathbf{1}\{c_{q,t}=o\},\quad o\in\mathcal O_q.
% \]
% Define the prior entropy $H_{q,i-1}=-\sum_{o\in\mathcal O_q}\mathbf p_{q,i-1}(o)\log \mathbf p_{q,i-1}(o)$ and per-round cross-modal hallucination $\mathrm{CMHR}_{q,i}\in[0,100]$ as defined in Section~\ref{sec:preliminary}.
% We exclude rounds $i=1$ (no prior distribution), questions with final $H=0$ (single-choice degeneracy), and rounds with missing scores.

\textbf{Random Guess Measure (RGM).}
For each question $q$ with option set $\mathcal O_q$ of size $n_q$, we collect $k_q$ independent generations with categorical choices $c_{q,i}$. The empirical prior before round $i\ge2$ is
\[
p_{q,i-1}(o) = \frac{1}{i-1}\sum_{t=1}^{i-1}\mathbf{1}[c_{q,t}=o], \quad o\in\mathcal O_q.
\]
We quantify dispersion via entropy
\[
H_{q,i-1} = -\sum_{o\in\mathcal O_q} p_{q,i-1}(o)\log p_{q,i-1}(o),
\]
and define question-level instability statistics (RGM and early-window RGM) based on aggregated entropy over rounds. Higher RGM indicates greater disagreement across rollouts and reflects unstable reasoning trajectories.

\textbf{Cross-Modal Hallucination Rate (CMHR).}
Each generation is evaluated by a panel of models $\mathcal M=\{\text{GPT-4o}, \text{Claude-Sonnet-4}, \text{Gemini}\}$ on:
truthfulness $T_{i,m}\in[1,5]$, informativeness $I_{i,m}\in[1,5]$, and consistency $C_{i,m}\in[0,1]$ (for VQA).
Aggregating across raters:
\[
\bar T_i=\tfrac{1}{|\mathcal M|}\sum_m T_{i,m},\quad\]
\[\bar I_i=\tfrac{1}{|\mathcal M|}\sum_m I_{i,m},\quad 
\bar C_i=\tfrac{1}{|\mathcal M|}\sum_m C_{i,m}.
\]
We define a reliability score
\[
\mathrm{Rel}_i=\Big(\tfrac{\bar T_i}{5}\Big)\Big(\tfrac{\bar I_i}{5}\Big)\bar C_i,
\]
and the cross-modal hallucination rate
\[
\mathrm{CMHR}=100\cdot\Big(1-\mathbb{E}_i[\mathrm{Rel}_i]\Big).
\]
Higher CMHR indicates poorer factual alignment or cross-modal inconsistency.

\textbf{Common preprocessing.}
(i) Compute $(H_{q,i-1},\,\mathrm{CMHR}_{q,i})$ for all rounds $i\ge2$.
(ii) Define \textbf{High--CMHR} as $\mathrm{CMHR}_{q,i}\ge\tau$, default $\tau$ at the global 75th percentile (threshold robustness below).
(iii) For each question, compute \textbf{churn} $S_q=\sum_{i=2}^{k_q}\mathbb{1}\{m_{q,i}\neq m_{q,i-1}\}$ where $m_{q,i}$ is the majority option under $\mathbf p_{q,i}$ (ties broken arbitrarily).

\textbf{Majority-path switching and churn.}
Let $m_{q,i}$ denote the majority option under $p_{q,i}$. A majority-path switch occurs when $m_{q,i}\neq m_{q,i-1}$. The total number of switches defines \emph{churn}:
\[
S_q=\sum_{i=2}^{k_q}\mathbf{1}\{m_{q,i}\neq m_{q,i-1}\}.
\]

\textbf{Evaluation protocol.}
We perform repeated CoT sampling ($k_q=15$) on $100$ medical questions (QA and VQA). We exclude the first round (no prior), degenerate cases with zero entropy, and missing evaluations. High-CMHR events are defined using a percentile-based threshold (default: 75th percentile). Statistical analyses (trend estimation, correlation, bootstrap confidence intervals, and mediation tests) follow the procedures detailed in Appendix~\ref{app:analysis-details}.

\begin{figure*}[!t]
  \includegraphics[width = 0.9\linewidth]{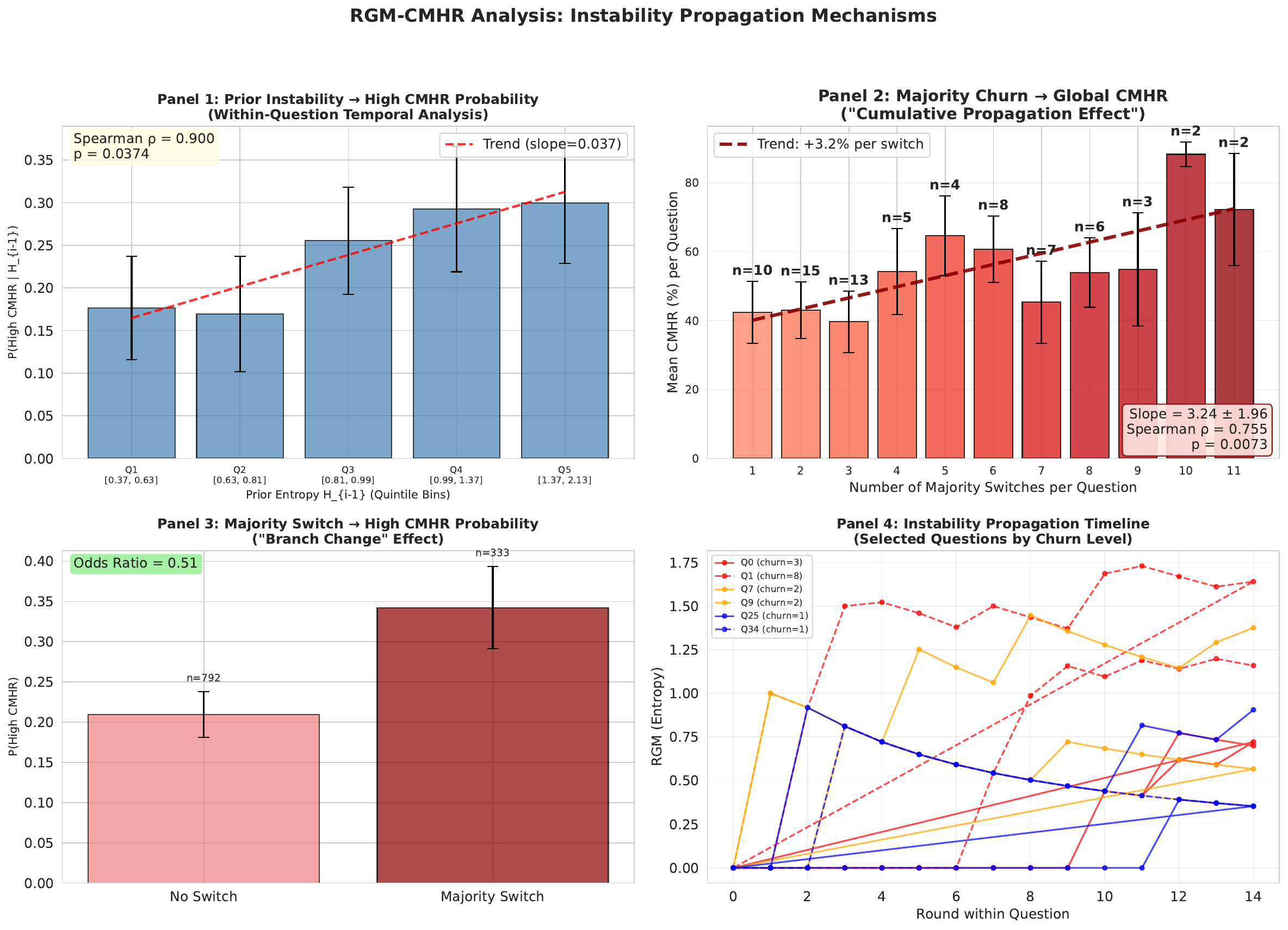} 
  \vspace{-6mm}
  \caption {
  Instability breeds hallucination.
    \emph{Panels 1–4:} Prior instability strongly predicts hallucination risk. 
    (1) Higher entropy at step $i-1$ increases the probability of high CMHR at step $i$. 
    (2) Majority-path churn accumulates globally and correlates with higher mean CMHR. 
    (3) A single majority-path switch produces an immediate +13.3\% increase in hallucination probability. 
    (4) Timelines show that high-churn questions propagate instability across reasoning steps, while low-churn questions remain stable.  
  }
  \vspace{-7mm}
  \label{fig:instability_mechanisms_appendix}
\end{figure*}

\subsection{Panel 1: Prior Instability $\to$ High CMHR (Within-Question Instantaneous Effect)}
\textbf{Goal.} Assess whether higher prior instability predicts hallucination risk in the subsequent generation.\\
\textbf{Construction.}
\begin{enumerate}[leftmargin=1.5em,itemsep=2pt]
\item Bin $H_{q,i-1}$ into quintiles (Q1--Q5) across all $(q,i)$.
\item For each bin $b$, estimate $\widehat{p}_b=\Pr(\mathrm{High\text{-}CMHR}_{q,i}=1\mid H_{q,i-1}\in b)$.
\item Plot $\widehat{p}_b$ with 95\% CIs from clustered bootstrap (5{,}000 resamples clustered by question).
\item Fit a monotone trend (OLS on bin midpoints) and report Spearman’s $\rho$ with exact $p$-values.
\end{enumerate}
\textbf{Note.} This is a within-question analysis using only prior entropy for the same item.

\subsection{Panel 2: Majority Churn $\to$ Global CMHR (Cumulative Propagation)}
\textbf{Goal.} Examine whether local instability aggregates into global inconsistency across the trajectory.\\
\textbf{Construction.}
\begin{enumerate}[leftmargin=1.5em,itemsep=2pt]
\item For each question $q$, compute churn $S_q$ and global inconsistency $\overline{\mathrm{CMHR}}_q=\frac{1}{k_q-1}\sum_{i=2}^{k_q}\mathrm{CMHR}_{q,i}$.
\item Plot $\overline{\mathrm{CMHR}}_q$ against $S_q$ (grouped if sparse) with mean$\pm$SE bars.
\item Estimate a robust OLS regression $\overline{\mathrm{CMHR}}_q=\alpha+\gamma S_q+\varepsilon_q$ (HC1 standard errors) and report $\hat\gamma$.
\item Report Spearman $\rho$ and $p$ for rank correlation.
\end{enumerate}

\subsection{Panel 3: Majority Switch $\to$ High CMHR (Branch-Change Effect)}
\textbf{Goal.} Translate entropy into a mechanism: test whether a branch switch increases hallucination risk within the same round.\\
\textbf{Construction.}
\begin{enumerate}[leftmargin=1.5em,itemsep=2pt]
\item Define a majority-switch indicator at round $i\ge2$: $Z_{q,i}=\mathbf{1}\{m_{q,i-1}\neq c_{q,i}\}$.
\item Form two groups: \textsf{no-switch} $(Z=0)$ vs. \textsf{switch} $(Z=1)$.
\item Estimate $\Pr(\mathrm{High\text{-}CMHR}\mid Z)$ with 95\% clustered bootstrap confidence intervals.
\item Report Fisher's exact test and the odds ratio (no-switch vs. switch).
\end{enumerate}
\textbf{Optional.} Logit with question fixed effects:
$\Pr(\mathrm{High\text{-}CMHR}_{q,i}=1)=\sigma(\alpha_q+\beta Z_{q,i})$; clustered SEs by question.

\subsection{Panel 4: Instability Propagation Timelines (Selected Questions)}
\textbf{Goal.} Visualize how instability accumulates over rounds for high- vs. low-churn items and link to the case tree.\\
\textbf{Construction.}
\begin{enumerate}[leftmargin=1.5em,itemsep=2pt]
\item Select representative questions at different churn levels (e.g., $S\in\{1,2,3,8,11\}$).
\item For each selected $q$, plot round-wise cumulative entropy $H_{q,i}$ (or $H_{q,i-1}$) over $i$; optionally smooth with a 3-point moving average.
\item Use a common y-axis and distinct line styles; annotate switches to show where branch changes occur.
\end{enumerate}
\textbf{Link.} Mark the high-entropy fork (Step~2) in the case-study tree; align rising segments in timelines with the error flow (misfocus/mis-mapping $\to$ wrong leaf).

\begin{figure*}[!ht]
% \vspace{-10pt}
\centering
\includegraphics[width=0.95\linewidth]{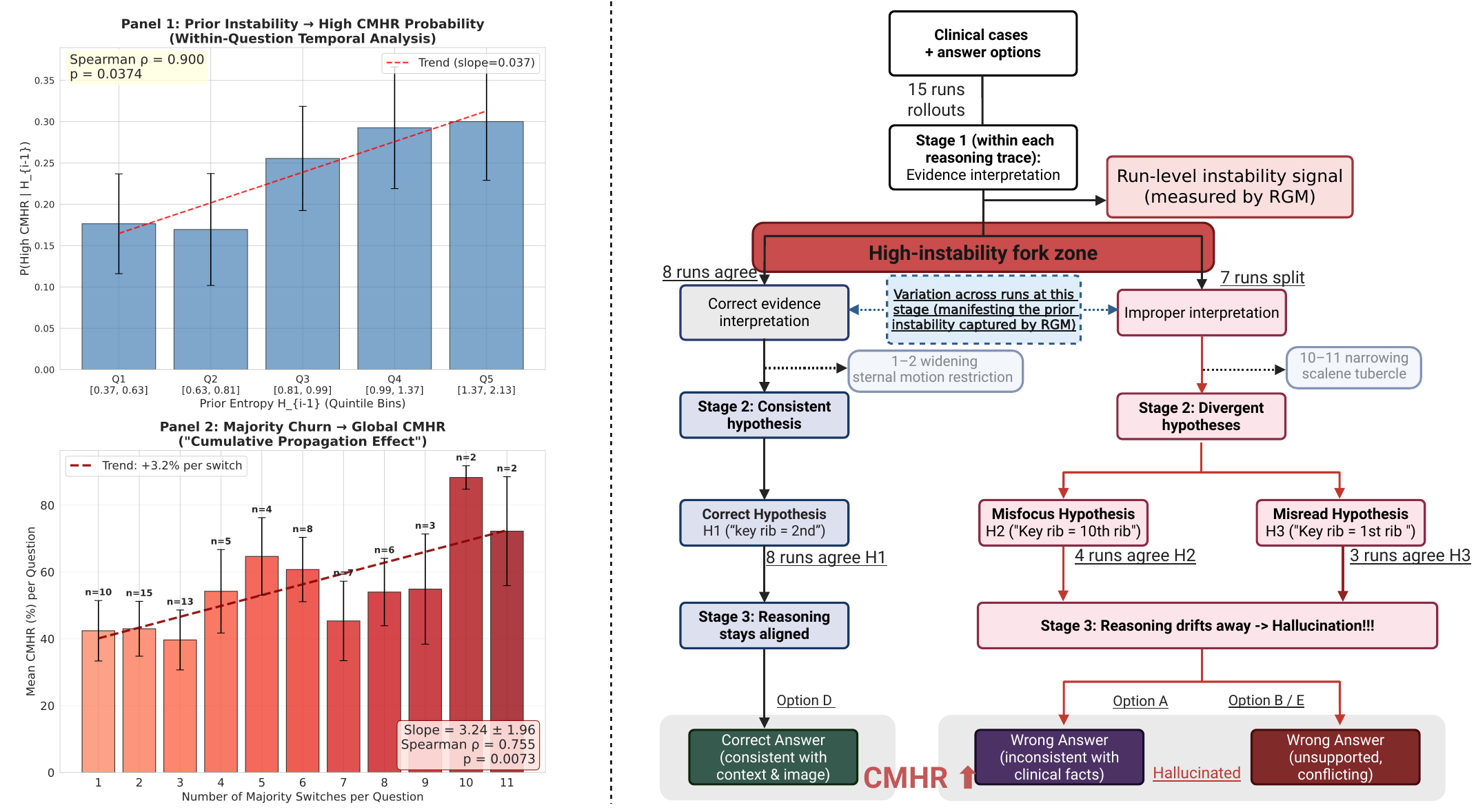}
\vspace{-2mm}
\caption{
% Instability breeds hallucination.
% \emph{Left (Panels 1–2):} Prior instability strongly predicts hallucination risk. 
% (1) Higher entropy at step $i-1$ increases the probability of high CMHR at step $i$. 
% (2) Majority-path churn accumulates globally and correlates with higher mean CMHR. 
% \emph{Right:} A multi-rollout case study illustrates how rollout-level prior instability 
% (measured by RGM) manifests within a single generation. At the evidence 
% interpretation stage, this prior cross-rollout instability makes it more likely to split into a high-instability fork (red). The unstable interpretations propagate into divergent hypotheses and eventually lead to hallucinated answers (red leaves), whereas the stable cluster of rollouts (blue/green) maintains a consistent hypothesis and arrives at the correct conclusion.
% \chen{I suggest moving this figure to appnendix, the layout is not pretty, and i think text only might be enough.}
Instability correlates with elevated hallucination risk.
\emph{Left (Panels 1–2)}: Higher prior-step entropy is associated with increased probability of high CMHR, and majority-path churn correlates with higher mean CMHR across questions.
\emph{Right}: A multi-rollout case study illustrates how run-level instability (measured by RGM) coincides with divergent evidence interpretation, with unstable forks leading to inconsistent conclusions while stable rollouts converge.
}
\label{fig:pilot-case}
\vspace{-4mm}
\end{figure*}

\textbf{Plotting and inference defaults.}
CIs: 95\% via clustered bootstrap (5{,}000 resamples, cluster=question) unless noted.

Nonparametrics: Spearman $\rho$ with exact (or large-sample) $p$.

Trends: OLS with HC1 robust SE; shaded band denotes 95\% CI.

Exclusions: rounds $i=1$, items with final $H=0$, and missing CMHR.

These analyses mirror real-world diagnostic uncertainty: early misinterpretation of evidence (instability) makes clinicians more likely to shift hypotheses, which increases the risk of introducing unsupported findings.

\textbf{Robustness variants.}
High--CMHR thresholds: $\tau$ at 70/75/80th percentiles or $z$-score $>0.5$.

Alternative binning (quintiles vs. deciles) yields qualitatively unchanged results.

Panel~1/3: fixed-effects logit; Panel~2: panel-OLS with question fixed effects.

Alternative switch definition: tie-aware majority and “top-2 margin $<\delta$” treated as \emph{uncertain} (robust to $\delta\in[0.05,0.15]$).

\textbf{Mediation analysis (early instability $\Rightarrow$ switching $\Rightarrow$ late hallucination).}
\emph{Causal chain and identification.}
We posit the directed chain
\[
\mathrm{RGM}^{\text{early}} \;(\,T\,) \;\longrightarrow\;
\mathrm{Switch} \;(\,M\,) \;\longrightarrow\;
\mathrm{CMHR}^{\text{late}} \;(\,Y\,),
\]
interpreted under: (A) \emph{Temporal ordering}: $T$ from early rounds, $M$ from a middle window, $Y$ from late rounds; (B) \emph{Sequential ignorability}: conditional on controls $X$ (item type, difficulty proxies, and question fixed effects), there are no unmeasured confounders of $T{\rightarrow}M$ and $M{\rightarrow}Y$ affected by $T$; (C) \emph{Consistency \& positivity}.\\
\emph{Specification.}
Let $T_i{=}\mathrm{RGM}^{\text{early}}$, $M_i{=}\mathrm{Switch}$ (count or rate in the mediator window), $Y_i{=}\mathrm{CMHR}^{\text{late}}$.
We estimate
\[
\begin{aligned}
\text{Mediator:}\quad & M_i=\alpha + a\,T_i + \gamma^\top X_i + \varepsilon_i,
\end{aligned}\]
\[
\begin{aligned}
\text{Outcome:}\quad & Y_i=\alpha' + c\,T_i + b\,M_i + \delta^\top X_i + \varepsilon'_i,
\end{aligned}
\]
with heteroskedasticity-robust SEs clustered at the question level; fixed-effects variants and nonparametric bootstrap (for $a$, $b$, and $ab$) are also considered.\\
\emph{Estimates.}
\[
\beta_a=5.319\!\pm\!1.360\ (p<10^{-4}),\quad\]
\[\beta_b=3.894\!\pm\!3.087\ (p=0.0158),\quad\]
\[\beta_c=9.201\!\pm\!18.927\ (p=0.3438).
\]
The indirect effect is significant (Sobel $z{=}2.353$, $p{=}0.0186$; $a\times b{=}20.712$), with a proportion mediated of $\sim$225\%, consistent with a suppression pattern where the mediator transmits predictive signal despite a weak marginal $T{\rightarrow}Y$ association.\\
\emph{Optional DAG.}

\begin{figure*}[!t]
\centering
\begin{tikzpicture}[
    >=Stealth,
    node distance=2.0cm and 2.8cm,
    box/.style={
        draw,
        rounded corners,
        align=center,
        minimum height=8mm,
        minimum width=26mm,
        inner xsep=5pt,
        inner ysep=4pt
    }
]

\node[box] (T) {$T=\mathrm{RGM}^{\text{early}}$};
\node[box, right=of T] (M) {$M=\mathrm{Switch}$};
\node[box, right=of M] (Y) {$Y=\mathrm{CMHR}^{\text{late}}$};
\node[box, above=1.0cm of M] (X) {$X$};

\draw[->] (T) -- (M);
\draw[->] (M) -- (Y);
\draw[->] (X) -- (M);
\draw[->] (X) -- (Y);

\draw[->]
  (T.north east)
  to[out=35, in=145]
  (Y.north west);

\end{tikzpicture}
\caption{Assumed mediation structure with controls $X$.}
\label{fig:mediation_structure}
\end{figure*}

\section{Dataset Statistics}
To better evaluate the performance of our model across diverse medical reasoning tasks, 
we collect and use several widely adopted datasets from both visual question answering (VQA) 
and text-based question answering (QA) domains. 
Table~\ref{tab:dataset_stat} summarizes the dataset statistics. 
Due to computational resource constraints, we report results on representative subsets for evaluation 
while ensuring coverage across different modalities (image + text vs. text-only). 
We select subsets by stratified sampling to preserve the diversity of question types and difficulty levels.
\label{sec:dataset_stat_appendix}

\begin{table}[!t]
    \centering
    \resizebox{\columnwidth}{!}{%
    \begin{tabular}{l c c c c}
        \toprule
        \textbf{Dataset} & \textbf{Task} & \textbf{Modality} & \textbf{Original size} & \textbf{Subset size} \\
        \midrule
        MedXpertQA-MM & VQA & Image + Text & 10,868 & 238 \\
        MedFrameQA & VQA & Image + Text &  2,851 & 200 \\
        PathVQA & VQA & Image + Text & 32,799 & 200 \\
        MedXpertQA-Text & QA & Text & 26,675 & 200 \\
        MedMCQA & QA & Text & 193,155 & 200 \\
        MedQA & QA & Text & 12,723 & 200 \\
        \bottomrule
    \end{tabular}%
    }
    \vspace{-3mm}
    \caption{Statistics of the six medical QA/VQA benchmarks used in our study. Each dataset includes the original size and the representative subset size used for evaluation, chosen to balance coverage and computational feasibility.}
    \vspace{-4mm}
    \label{tab:dataset_stat}
\end{table}

\section{Algorithmic Details}
The complete MedMMV workflow is formalized in Algorithm~\ref{alg:medmmv_framework}. The procedure begins with Stage 1 (Hypothesis Generation), where the MLLM produces $k$ short diagnostic rollouts from clinical notes and images, yielding a diverse set of initial reasoning paths $\mathcal{P}_{\text{init}}$.

In Stage 2 (Parallel Evidence-Grounded Refinement), we first construct a unified evidence graph $\mathcal{E}$ using the three specialist modules: \texttt{TextDoctor} for structured entities from notes, \texttt{ImageDoctor} for image-derived findings, and \texttt{WebSearch} for external clinical knowledge. Each initial path is then refined in parallel under the supervision of the Hallucination and Consistency Detector (HD Supervisor). At each iteration, reasoning claims are fact-checked against $\mathcal{E}$; if inconsistencies or unsupported statements are detected, targeted feedback triggers the \texttt{AutoRepair} process, revising the path until convergence. The outcome is a set of refined paths paired with their supporting subgraphs $\mathcal{C}_{\text{refined}}$.

Finally, Stage 3 (Decision Aggregation) selects the optimal diagnosis. Each candidate path is evaluated by the Combined Uncertainty (CU) Scorer, which integrates evidence alignment, coherence, and repair cost into a single reliability score. The highest-scoring path $(\hat{p}, \mathcal{E}_{\hat{p}})$ is chosen, and the final diagnosis $\hat{y}$ is extracted from it. This ensures that the reported output is not only accurate but also explicitly verified against multimodal evidence.

\begin{algorithm*}[!t]
\caption{The MedMMV Framework Algorithm}
\label{alg:medmmv_framework}
\SetKwComment{tcp}{// }{}
\SetKwFunction{FactCheck}{FactCheck}
\SetKwFunction{HDSupervisor}{HDSupervisor}
\SetKwFunction{AutoRepair}{AutoRepair}
\SetKwFunction{TextDoctor}{TextDoctor}
\SetKwFunction{ImageDoctor}{ImageDoctor}
\SetKwFunction{WebSearch}{WebSearch}
\SetKwFunction{BuildEvidenceGraph}{BuildEvidenceGraph}
\SetKwFunction{CUScorer}{CUScorer}
\SetKwFunction{ExtractDiagnosis}{ExtractDiagnosis}

\KwIn{Clinical notes $\mathcal{T}$; medical images $\mathcal{I}$; number of rollouts $k$.}
\KwOut{Final diagnosis $\hat{y}$; diagnostic text $\hat{p}$; evidence graph $\mathcal{E}_{\hat{p}}$.}

\BlankLine

\begin{minipage}[t]{0.48\textwidth}
\small

\tcp{Stage 1: Diverse initial hypotheses}
$\mathcal{P}_{\mathrm{init}} \leftarrow \emptyset$\;
\For{$i \leftarrow 1$ \KwTo $k$}{
  $p^{(i)}_{\mathrm{init}}
  \leftarrow
  \mathrm{MLLM.generate\_rollout}(\mathcal{T}, \mathcal{I})$\;
  $\mathcal{P}_{\mathrm{init}}
  \leftarrow
  \mathcal{P}_{\mathrm{init}} \cup \{p^{(i)}_{\mathrm{init}}\}$\;
}

\BlankLine
\tcp{Stage 2: Build evidence graph}
$\mathcal{E} \leftarrow
\BuildEvidenceGraph(
  \TextDoctor(\mathcal{T}),
  \ImageDoctor(\mathcal{I}),
  \WebSearch(\cdots)
)$\;

$\mathcal{C}_{\mathrm{refined}} \leftarrow \emptyset$\;

\BlankLine
\tcp{Initialize parallel refinement}
Each $p_{\mathrm{init}} \in \mathcal{P}_{\mathrm{init}}$ is refined independently using the shared evidence graph $\mathcal{E}$\;

\end{minipage}
\hfill
\begin{minipage}[t]{0.48\textwidth}
\small

\tcp{Parallel refinement of each initial path}
\For{$p_{\mathrm{init}}$ \textbf{in parallel} from $\mathcal{P}_{\mathrm{init}}$}{
  $p_{\mathrm{current}} \leftarrow p_{\mathrm{init}}$\;

  \While{not converged}{
    $\mathrm{feedback}
    \leftarrow
    \HDSupervisor(
      \FactCheck(p_{\mathrm{current}}, \mathcal{E})
    )$\;

    \If{$\mathrm{feedback} = \emptyset$}{
      \textbf{break}\;
    }

    $p_{\mathrm{current}}
    \leftarrow
    \AutoRepair(p_{\mathrm{current}}, \mathrm{feedback})$\;
  }

  $p_{\mathrm{final}} \leftarrow p_{\mathrm{current}}$\;

  $\mathcal{E}_{p}
  \leftarrow
  \mathrm{ExtractRelevantSubgraph}
  (\mathcal{E}, p_{\mathrm{final}})$\;

  $\mathcal{C}_{\mathrm{refined}}
  \leftarrow
  \mathcal{C}_{\mathrm{refined}}
  \cup
  \{(p_{\mathrm{final}}, \mathcal{E}_{p})\}$\;
}

\BlankLine
\tcp{Stage 3: Aggregation and final selection}
$(\hat{p}, \mathcal{E}_{\hat{p}})
\leftarrow
\arg\max\limits_{
  (p_j, \mathcal{E}_j)
  \in
  \mathcal{C}_{\mathrm{refined}}
}
\CUScorer(p_j, \mathcal{E}_j)$\;

$\hat{y}
\leftarrow
\ExtractDiagnosis(\hat{p})$\;

\BlankLine
\Return $\hat{y}, \hat{p}, \mathcal{E}_{\hat{p}}$\;

\end{minipage}

\end{algorithm*}

\section{Prompt and Output Example of Agents}
\label{app:prompts_app}
To ensure transparency and reproducibility, we provide representative prompts and output examples of the core agents in MedMMV. These include the \emph{Text Doctor}, which extracts structured findings from clinical notes, the \emph{Image Doctor}, which objectively describes medical imaging features, and the \emph{Hallucination Detector}, which flags fabricated or unsupported reasoning. We also illustrate the \emph{Researcher} module’s evidence-graph–based search compared to whole-context search, highlighting how structured verification enables more precise and traceable reasoning. Together, these examples demonstrate how MedMMV operationalizes controllable reasoning through standardized agent behaviors.

\subsection{Text Doctor Prompt}
\begin{tcolorbox}[
    colback=lightbluebg!30!white,
    colframe=blueframe,
    title=Expert Agent Prompt,
    fonttitle=\bfseries\large
]
PROMPT =  """

You are a medical domain expert analyzing clinical text data. Your task is to identify key symptoms, conditions, and clinical findings from the patient information provided.

CLINICAL TEXT:
\{\textbf{Clinical text}\}

Consider this may be a medical exam question with a specific diagnosis in mind. Extract family medical history, past medical history, relevant symptoms, medical conditions, and clinical findings carefully. Organize findings by body system or symptom category.
Note any significant patient history that could impact diagnosis.

FORMAT YOUR RESPONSE EXACTLY AS FOLLOWS:

FAMILY MEDICAL HISTORY:

- [ family condition 1 ]

PAST MEDICAL HISTORY:
- [ past condition 1 ]

EXTRACTED SYMPTOMS:

- [symptom 1]

CLINICAL FINDINGS:

[Body System 1]:

- [finding 1]

[Body System 2]:

- [finding 1]

Ensure your output follows this exact format for automated processing.
"""
\end{tcolorbox}

\subsection{Image Doctor Prompt}
\begin{tcblisting}{
    enhanced,
    breakable,
    colback=lightbluebg!30!white,
    colframe=blueframe,
    title=Expert Agent Prompt,
    fonttitle=\bfseries\large,
    listing only,
    listing options={
        basicstyle=\ttfamily\small,
        breaklines=true,
        breakatwhitespace=false,
        columns=fullflexible,
        keepspaces=true,
        showstringspaces=false
    },
    before upper={\setlength{\parindent}{0pt}},
}
PROMPT =  """

Your task is to carefully and objectively describe the visible imaging findings, without making diagnostic judgments or clinical assumptions.

PATIENT INFORMATION:
{patient info}

MEDICAL IMAGES:
{medical images}

Please identify and describe all visible abnormalities, focusing only on what is directly observable. Do not speculate or provide diagnoses.

For each finding, describe the following:

- Region or organ involved

- Distribution: focal / diffuse / multifocal

- Number: single / few / multiple / innumerable

- Size: provide approximate range (e.g., 1-3 mm, <1 cm, >3 cm)

- Density: solid / ground-glass / cavitary / mixed

- Margins: well-defined / ill-defined

- Laterality: unilateral / bilateral

- Associated features: pleural effusion, lymphadenopathy, airway distortion, cavitation, etc. (only if visible)

Be concise, objective, and specific. Avoid speculative language or uncertain modifiers (e.g., "may represent", "possibly"). Only report what is visually evident.

FORMAT YOUR RESPONSE EXACTLY AS FOLLOWS:

IMAGE FINDINGS:

- [Region/Organ]&#58; [Concise structured description with above attributes]
"""
\end{tcblisting}
\subsection{Researcher Pseudocode and Comparing with Whole Context Search}
\label{app:search_case_compare_app}

We adopt an evidence-graph approach instead of whole-context search, which often reduces retrieved text to undifferentiated facts and loses critical relationships among findings, interventions, and outcomes. By explicitly modeling entities and typed links, and verifying with external literature, evidence graphs enable clinically meaningful connections, conflict detection, and traceable reasoning.

\begin{algorithm}[t]
\caption{Medical Literature Web Search}
\label{alg:web_search}
\SetKwComment{tcp}{// }{}
\SetKwFunction{GetObservations}{GetObservations}
\SetKwFunction{EmptySearchResult}{EmptySearchResult}
\SetKwFunction{GeneratePairs}{GeneratePairs}
\SetKwFunction{InitializeChatGPT}{InitializeChatGPT}
\SetKwFunction{BuildMedicalQuery}{BuildMedicalQuery}
\SetKwFunction{WebSearchAPI}{WebSearchAPI}
\SetKwFunction{ExtractContent}{ExtractContent}
\SetKwFunction{FormatFindings}{FormatFindings}
\SetKwFunction{SearchState}{SearchState}

\KwIn{Search state $state$; configuration $config$.}
\KwOut{Web search results with medical findings.}

\BlankLine
\tcp{Stage 1: Retrieve observations}
$evidencegraph \leftarrow \GetObservations(state)$\;
\If{$evidencegraph = \emptyset$}{
    \Return \EmptySearchResult()\;
}

\BlankLine
\tcp{Stage 2: Initialize search candidates}
$combinations \leftarrow \GeneratePairs(evidencegraph, \text{max}=8)$\;
$llm \leftarrow \InitializeChatGPT(config)$\;
$findings \leftarrow [\,]$\;

\BlankLine
\tcp{Stage 3: Query medical literature}
\For{$combo \in combinations$}{
    $query \leftarrow \BuildMedicalQuery(combo)$\;
    $results \leftarrow \WebSearchAPI(query)$\;
    $context \leftarrow \ExtractContent(results)$\;
    $analysis \leftarrow llm.\text{Analyze}(context)$\;
    $findings.\text{append}(combo, analysis)$\;
}

\BlankLine
\tcp{Stage 4: Merge findings}
$merged \leftarrow \FormatFindings(findings)$\;
\Return \SearchState($merged, combinations$)\;

\end{algorithm}

\begin{tcblisting}{
    enhanced,
    breakable,
    colback=lightbluebg!30!white,
    colframe=blueframe,
    title=Whole Context Search,
    fonttitle=\bfseries\large,
    listing only,
    listing options={
        basicstyle=\ttfamily\small,
        breaklines=true,
        breakatwhitespace=false,
        columns=fullflexible,
        keepspaces=true,
        showstringspaces=false
    }
}
RESULTS =  """

Page 1:
Purpose
To define the role of MRI in the diagnosis and management of Chance-type flexion distraction injury.

Results
At MRI, combined bony and soft tissue injuries were more common than either bone or soft tissue damage alone, and occurred at the thoracolumbar junction primarily. Contiguous vertebral injury was seen in 18 cases, with non-contiguous injury in 7 cases. Posterior ligamentous complex disruption occurred in all cases. Extensive subcutaneous and para-spinal muscle oedema was seen in all cases extending over several segments. Horizontally orientated fractures of the posterior neural arches produced a distinctive MRI pattern -"Sandwich sign"- consisting of linear haemorrhage framed by marrow oedema. Extension of the fracture into the posterior vertebral body outline occurred in 3 cases, with fracture displacement into the canal. The posterior vertebral body height remained unchanged or increased in these 3 cases.

Conclusion:
The MR features of CTFD injuries are recognisable and reproducible, allowing a confident differentiation from "unstable burst injuries".

Page 2:
Context:
Arthroscopic reconstruction of ACL is an effective method to restore knee stability after ACL rupture. Postoperative septic arthritis (SA) is very uncommon while the incidence of serious complications range between 0.14% and 1.8%. Some of the devastating consequences of septic arthritis can encompass hyaline cartilage damage, arthrofibrosis, and in rare cases amputation. The purpose of this study was to evaluate the effect of gentamicin irrigation solutions as a process to restrain septic arthritis following arthroscopic ACL reconstruction.

Results:
Seven patients developed SA, four of whom were from SALINE group (2.2%) and three from Gentamycin group (0.23%). The incidence rate of SA after arthroscopic ACL reconstruction was significantly lower (P <0.05) when irrigated with gentamicin solution than merely with saline solution.

Conclusion:
Gentamicin irrigation solution has a preservative and protective effect against SA development following arthroscopic ACL reconstruction. We recommend evaluating this technique as a way in order to depreciate the prevalence of SA after ACL reconstruction.
"""
\end{tcblisting}

\begin{tcolorbox}[
    enhanced,
    breakable,
    parbox=false,
    colback=lightbluebg!30!white,
    colframe=blueframe,
    title=Evidence Graph Search,
    fonttitle=\bfseries\large,
    before upper={\small}
]
RESULTS =  ``''

\medskip

\textbf{[Fever + Pain]}: Fever and pain are often associated, with pain sometimes being a symptom of the underlying illness causing the fever, or the fever itself causing symptoms like headache or muscle aches. In children presenting with fever and limb pain, acute lymphoblastic leukemia (ALL) should be considered, especially if accompanied by fatigue, bruising, or bleeding. In infants and toddlers, ibuprofen may be more effective than acetaminophen for fever reduction and pain relief.

\medskip

\textbf{[Fever + Wound drainage at the surgical site]}: Fever and wound drainage at the surgical site can indicate a surgical site infection (SSI), which may involve superficial or deep tissues and sometimes manifests with heat, redness, swelling, and purulent exudate. While fever is a common response to surgery, a persistent or high fever accompanied by wound drainage should be evaluated to rule out infection.

\medskip

\textbf{[Fever + ruptured Achilles tendon with a 5 cm gap]}: While fever is not directly established as a common association of a ruptured Achilles tendon with a 5 cm gap, systemic diseases including infections can be related to Achilles tendon injuries. A ruptured Achilles tendon is often characterized by sudden pain, swelling, and impaired movement. A larger gap size in the ruptured tendon may predict lower patient-reported outcomes.

\medskip

\textbf{[Fever + Elevated ESR of 29 mm/hr (normal range: 0)]}: Fever with an elevated ESR of 29 mm/hr (normal range: 0) can be associated with systemic or bone infections, heart conditions, rheumatic fever, severe skin infections, tuberculosis, autoimmune disorders, certain cancers like lymphoma or multiple myeloma, kidney or thyroid disease, anemia, pregnancy, diabetes mellitus, end-stage renal disease, heart disease, malignancy, allergic vasculitis...

\medskip

\textbf{[Fever + Past Medical History: [Achilles tendon repair]]}: A fever following Achilles tendon repair could indicate a potential infection, warranting immediate contact with the referring physician.

\medskip

``''
\end{tcolorbox}

% \subsection{Example of Supervised Path Self-refinement}
\subsection{Hallucination Detector Prompt}

\begin{tcolorbox}[
    colback=lightbluebg!30!white,
    colframe=blueframe,
    title=Expert Agent Prompt,
    fonttitle=\bfseries\large
]
PROMPT =  """

You are a medical hallucination detector. Your task is to identify any fabricated, imagined, or unsupported medical information in the analysis.

Original Clinical Text
{\textbf{original clinical text}}

Analysis Type: \{\textbf{analysis type}\}

Analysis to Check
\{\textbf{analysis text}\}

Please identify:

1. FABRICATED INFORMATION

2. UNSUPPORTED CLAIMS

3. CONTRADICTIONS

Format your response as:

HALLUCINATION DETECTED: [YES/NO]

FABRICATED INFORMATION:
- [List any fabricated details, or "None detected"]

UNSUPPORTED CLAIMS:
- [List any unsupported medical claims, or "None detected"]

CONTRADICTIONS:
- [List any contradictions, or "None detected"]

CONFIDENCE SCORE: [0-100]\%
(How confident you are in this hallucination assessment)

RECOMMENDATION:
Provide specific actionable recommendation 

- choose from: ACCEPT AS IS, ACCEPT WITH CAUTION, REVISE REDUCE FABRICATION, REVISE STRENGTHEN EVIDENCE, REVISE REMOVE CONTRADICTIONS, REVISE COMPREHENSIVE, REJECT HIGH HALLUCINATION, REJECT UNRELIABLE

SPECIFIC ACTIONS:
List 2 to 3 specific actions to improve the analysis.

SEVERITY: LOW/MEDIUM/HIGH
"""
\end{tcolorbox}

The example JSON output of hallucination detector is as follows:

\begin{Verbatim}[breaklines=true, breakanywhere=true, fontsize=\small]
{
  "analysis_name": "Image Doctor Analysis",
  "attempt": 1,
  "hallucination_detected": "YES",
  "recommendation": "REVISE_REDUCE_FABRICATION",
  "total_issues": 4,
  "severity": "MEDIUM",
  "improvement_guidance_applied": false,
  "improvement_instructions": [
    "Action: Base imaging interpretation only on actually visible findings in the image.",
    "Action: Eliminate unsupported claims about the number of lesions.",
    "Do not fabricate: Lesions vary, approximately 1–3 cm in diameter is not mentioned in the original clinical text.",
    "Unsupported claim to avoid: The description of multiple lesions is not explicitly supported by the original clinical text or visible in the image.",
  ],
  "previous_warnings_count": 0
}
\end{Verbatim}

\section{Case Study}
To qualitatively demonstrate the capabilities of MedMMV, we present several case studies in the Appendix~\ref{sec:detailed_results_appendix} and Figure~\ref{fig:case_study_fig_appendix}. We compare the reasoning process of our agent with that of a direct CoT. 
These case studies illustrate two important properties. First, in \emph{ambiguous scenarios} where multiple diagnostic hypotheses are initially plausible, MedMMV expands paths in parallel and uses the evidence graph to explicitly weigh supporting vs.\ conflicting findings. This prevents premature commitment to a single hypothesis and allows the system to maintain clinically reasonable alternatives until sufficient evidence is gathered. These case studies illustrate two important properties. First, in \emph{ambiguous scenarios} where multiple diagnostic hypotheses are initially plausible, MedMMV expands paths in parallel and uses the evidence graph to explicitly weigh supporting vs.\ conflicting findings. This prevents premature commitment to a single hypothesis and allows the system to maintain clinically reasonable alternatives until sufficient evidence is gathered. Second, when \emph{errors arise}, such as fabricated or misinterpreted evidence in an early path, the hallucination detector flags the inconsistency and feeds it back to the corresponding reasoning block. This allows MedMMV to actively revise the erroneous evidence and update the evidence graph. The case studies highlight how MedMMV’s controllable reasoning enables it to handle ambiguous clinical scenarios, correct its own errors, and ultimately lead to a more reliable and clinically sound diagnosis.

\subsection{Typical Repaired Example Comparison.}
\label{sec:detailed_results_appendix}
To further highlight the advantages of MedMMV over direct chain-of-thought (CoT) reasoning, we present a representative example where the baseline model fails while our approach successfully repairs the reasoning (Table~\ref{tab:case_study_outputs}). The case involves a patient with an infected re-ruptured Achilles tendon. Direct CoT incorrectly prioritizes immediate reconstruction, overlooking the contraindication posed by active infection. In contrast, MedMMV explicitly aligns textual, visual, and clinical evidence to recognize infection as a dominant factor, thereby deferring reconstruction in favor of thorough debridement. This example illustrates how MedMMV prevents reasoning shortcuts, repairs erroneous diagnostic chains, and produces more clinically faithful outputs by enforcing evidence alignment and highlighting bifurcation points during the reasoning process.

\begin{table*}[!ht]
\centering
  % \vspace{-7mm}
\caption{Comparing the Outputs of Direct CoT and MedMMV on GPT-oss-120B.}
\label{tab:case_study_outputs}
\begin{tabular}{p{\textwidth}}
\toprule
\textbf{Question} \\
\midrule
A 36-year-old male patient presents with fever, pain, and wound drainage at the surgical site 4 months following an Achilles tendon repair. Figure A shows the clinical presentation. Laboratory testing reveals an elevated ESR of 29 mm/hr (normal range: 0-22 mm/hr). During the planned debridement and irrigation procedure, it is discovered that the Achilles tendon has re-ruptured with a 5 cm gap. Given these findings, what is the most appropriate next step in surgical management? 

\textbf{Options:} 
A: Primary repair of the re-ruptured Achilles tendon

B: Repair of the re-ruptured Achilles tendon with a turndown procedure

C: Repair of defect with flexor hallucis longus tendon transfer

D: Debridement of necrotic and infected tendon tissue, with no attempt at reconstruction

E: V-Y plasty of the re-ruptured Achilles tendon \\
% \bottomrule
\end{tabular}

% \vspace{0.5em}

% Comparison section - two columns
\begin{tabular}{p{0.48\textwidth}p{0.48\textwidth}}
\toprule
\textbf{Direct CoT} & \textbf{MedMMV (ours)} \\
\midrule
\textbf{Selected Answer: (C)}  

\textbf{Evidence cited:}  

-- \textcolor{red}{Large gap (5 cm) $\Rightarrow$ requires FHL transfer immediately.}  

-- \textcolor{red}{Infection can be managed simultaneously with reconstruction.}  

\textbf{Issue:} \textcolor{red}{Fails to recognize that active infection contraindicates immediate reconstruction.}  

\textbf{Clinical Reasoning:} The infection and large gap necessitate both adequate debridement and effective reconstruction. Flexor hallucis longus tendon transfer is appropriate for large defects and aids in eliminating infected tissue while providing functional reconstruction. 
& 
\textbf{Selected Answer: (D)}  

\textbf{Corrected Evidence:}  

-- \textcolor{blue}{Active infection + necrotic tissue $\Rightarrow$ reconstruction must be deferred.}  

-- \textcolor{blue}{Primary repair (A), turndown (B), V-Y plasty (E) are contraindicated during infection.}  

-- \textcolor{blue}{FHL transfer (C) is valid for chronic >3 cm defects, but only after infection eradication.}  

\textbf{Reasoning:} \textcolor{blue}{Immediate priority is thorough debridement of infected tissue to prepare for later reconstruction.}  

\textbf{Clinical Reasoning:} Given these findings, at this stage the priority is infection control rather than reconstruction. Debridement, which involves debridement of necrotic and infected tissue without reconstruction, is a more conservative approach that addresses the infection first. This is crucial to prevent further complications and to create a healthier environment for any future reconstructive procedures. Therefore, the most appropriate next step in surgical management, given the active infection and necrotic tissue, is to focus on debridement. \\
\bottomrule
\end{tabular}
\vspace{-7mm}
\end{table*}

\subsection{Example with Intermediate Output.}
\label{sec:case_study_appendix}
\begin{figure*}[htbp]
  \includegraphics[width = \linewidth]{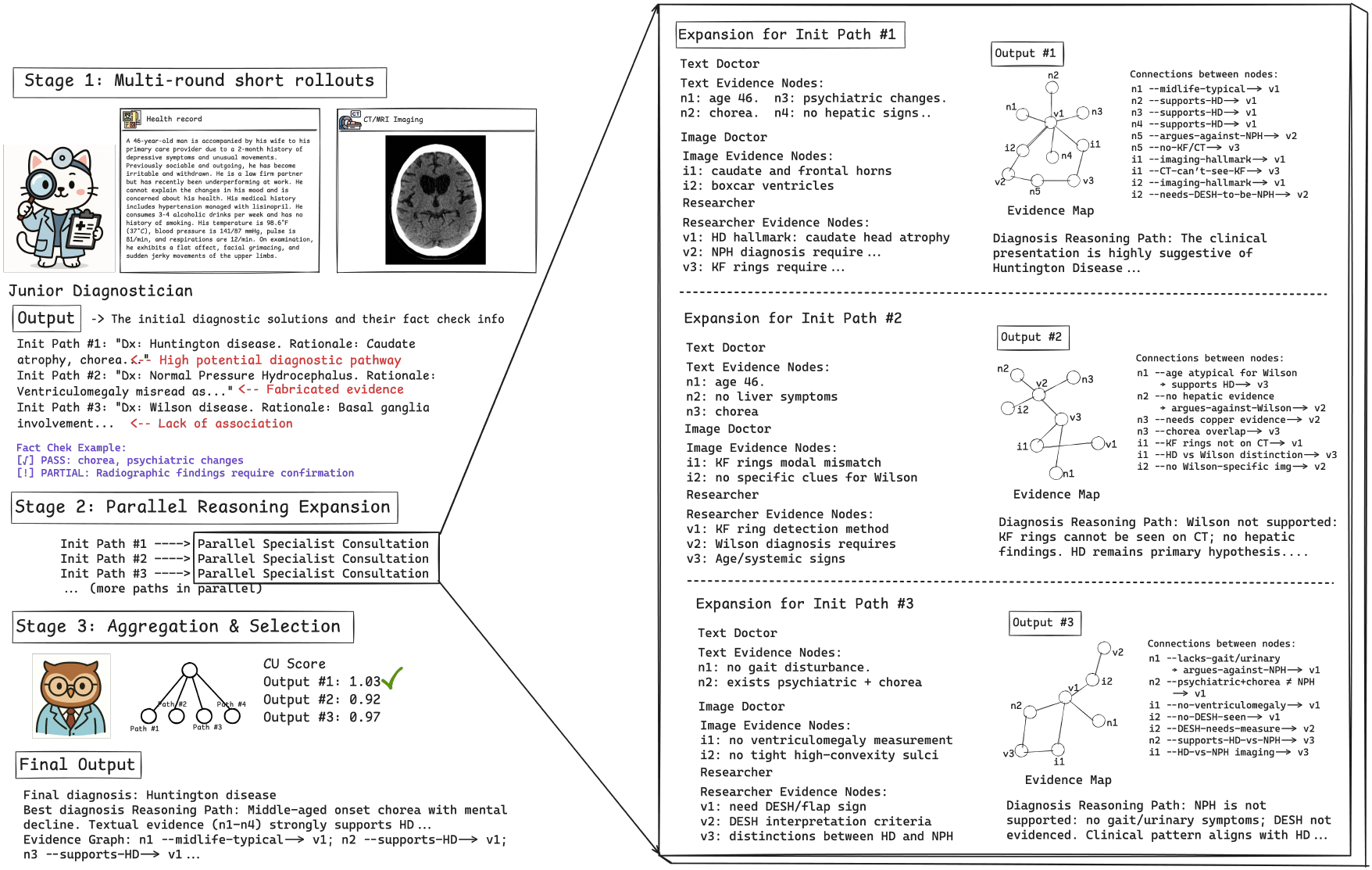} 
  \caption {Case study example of MedMMV with intermediate outputs.
  }
    \vspace{-6mm}
  \label{fig:case_study_fig_appendix}
\end{figure*}

To better illustrate how MedMMV performs step-by-step reasoning, we provide a representative case study with intermediate outputs (Figure~\ref{fig:case_study_fig_appendix}). The system begins with \textbf{Stage 1}, where multiple short diagnostic rollouts are generated by a junior diagnostician, each associated with an initial hypothesis and a preliminary fact check. In \textbf{Stage 2}, these candidate diagnostic paths are expanded in parallel through consultation with text-based doctors, image specialists, and researcher modules. Each expansion produces structured evidence nodes and cross-modal connections, resulting in evidence maps that make the reasoning process transparent. Finally, in \textbf{Stage 3}, the outputs from parallel paths are aggregated and scored. The system then selects the most consistent and well-supported reasoning chain as the final output. This case demonstrates how MedMMV integrates textual, visual, and clinical knowledge sources to arrive at a robust diagnosis while explicitly showing intermediate reasoning states.

\section{Ablation Study Details}
\subsection{Ablation Settings}
\label{sec:detailed_ablation_appendix}
This section provides precise description of the ablation settings presented in Table~\ref{tab:ablation_medqa}.

\textbf{Full model.}
This is the complete MedMMV system as described in the main paper. It utilizes all stages. This configuration serves as the main baseline for all comparisons.

\textbf{w/o parallel reasoning expansion.}
In this setting, we replace the core architecture of generating and expanding multiple parallel paths. Instead, the model employs a single, sequential reasoning process with evidence-grounded reasoning.
The model iteratively refines its single line of reasoning, but does not explore the diverse hypothesis space that our parallel framework is designed to cover. This tests the benefit of concurrent exploration.

\textbf{w/o HD supervisor.}
The Hallucination Detection (HD) Supervisor, a key component within each Reasoning Expansion Module, is deactivated. Consequently, the reasoning paths are generated without the internal self-correction loop. The Auto-Repair step, which relies on feedback from the supervisor to fix logical fallacies or factual errors, is effectively skipped. This ablation isolates the impact of our intra-path consistency and factuality checks.

\textbf{w/o TextDoctor agent.}
The system is prohibited from invoking the TextDoctor agent during the Evidence Integration step of Stage 2. All reasoning must be based on evidence sourced from the original text input. This directly measures the contribution of structured text evidence extraction from clinical information.

\textbf{w/o ImageDoctor agent.}
Similar to the TextDoctor ablation, the ImageDoctor agent is made unavailable. Forcing the system to rely solely on original medical image information rather than MedMMV's refined and grounded visual evidence. This configuration is only applicable to the multimodal MedXpert-MM task and is designed to quantify the value of visual grounding.

\textbf{w/o WebSearch agent.}
The WebSearch agent is disabled, preventing the model from accessing external knowledge sources to verify facts or gather additional context. The reasoning process is confined to the information explicitly provided in the clinical notes and images. This setup assesses the importance of external, up-to-date knowledge in complex medical reasoning.

\textbf{Random decision instead of CU scoring.}
In the final Aggregation \& Selection stage (Stage 3), the CU Scorer is bypassed. After the collection of all refined paths $\{ (p_{\text{final}}, \mathcal{E}_p) \}$, the final output is selected by choosing one path uniformly at random, rather than selecting the path with the highest confidence and utility score. This ablation directly tests the efficacy of our uncertainty-aware selection mechanism against a naive baseline.

\subsection{More Abaltion Study Results}
\label{app:full_abaltion}
Except for MedXpert-MM and MedXpert-Text, we also conduct abaltion experiment on the other four datasets.

\begin{table*}[!ht]
\vspace{-2mm}
\centering
\caption{Ablation study on all six datasets. \textit{\textbf{Note:}} Acc: Accuracy (\%); T: Truthfulness (1--5); I: Informativeness (1--5); T$\times$I: normalized product in [0,100], computed as $(T \times I)/25 \times 100$. ImageDoctor ablations are not applicable (--) for text-only tasks. Bold indicates full MedMMV with best performance. Each dataset samples 200 cases.}
\captionsetup{justification=raggedright,singlelinecheck=false,font=tiny}
\vspace{-2mm}
\label{tab:full_ablation_medqa}

\setlength{\tabcolsep}{2pt}
\scriptsize
\resizebox{\textwidth}{!}{%
\begin{tabular}{@{}ll*{12}{c}@{}}
\toprule
\multirow{3}{*}{\textbf{Category}} & \multirow{3}{*}{\textbf{Methods}} & \multicolumn{4}{c}{\textbf{MedXpert-MM}~(\TextCircle\ImageCircle)} & \multicolumn{4}{c}{\textbf{MedFrameQA}~(\TextCircle\ImageCircle)} & \multicolumn{4}{c}{\textbf{PathVQA}~(\TextCircle\ImageCircle)} \\
\cmidrule(lr){3-6} \cmidrule(lr){7-10} \cmidrule(lr){11-14}
& & Acc & T & I & T$\times$I & Acc & T & I & T$\times$I & Acc & T & I & T$\times$I \\
\midrule
\multirow{1}{*}{Full model}
& \textbf{MedMMV (GPT-5)} & \textbf{73.58} & \textbf{4.26} & \textbf{4.73} & \textbf{82.89} & \textbf{46.95} & \textbf{3.44} & \textbf{4.57} & \textbf{65.84} & \textbf{67.91} & \textbf{3.51} & \textbf{4.59} & \textbf{68.00} \\
\hdashline
\multirow{2}{*}{Core framework}
& w/o Path Expansion with multi-round short CoT & 71.45 & 4.23 & 4.70 & 79.52 & 43.28 & 3.18 & 4.32 & 61.15 & 64.13 & 3.26 & 4.38 & 63.47 \\
& w/o Self-feedback halluc. detector & 69.80 & 4.19 & 4.68 & 76.45 & 41.05 & 2.95 & 4.41 & 58.92 & 61.74 & 3.08 & 4.45 & 61.28 \\
\hdashline
\multirow{3}{*}{Expert agents}
& w/o TextDoctor agent & 72.10 & 4.24 & 4.71 & 80.15 & 44.12 & 3.26 & 4.38 & 62.47 & 65.27 & 3.35 & 4.43 & 64.85 \\
& w/o ImageDoctor agent & 72.95 & 4.25 & 4.72 & 81.60 & 45.36 & 3.35 & 4.49 & 64.08 & 66.54 & 3.43 & 4.52 & 66.39 \\
& w/o WebSearch agent & 73.15 & 4.25 & 4.72 & 82.05 & 46.17 & 3.39 & 4.53 & 65.13 & 67.18 & 3.47 & 4.56 & 67.24 \\
\hdashline
\multirow{1}{*}{Decision strategy}
& Random decision instead of CU scoring & 68.55 & 4.18 & 4.65 & 73.90 & 38.47 & 2.64 & 4.15 & 54.73 & 58.26 & 2.79 & 4.23 & 56.41 \\
\midrule
\multirow{3}{*}{\textbf{Category}} & \multirow{3}{*}{\textbf{Methods}} & \multicolumn{4}{c}{\textbf{MedXpert-Text}~(\TextCircle)} & \multicolumn{4}{c}{\textbf{MedQA}~(\TextCircle)} & \multicolumn{4}{c}{\textbf{MedMCQA}~(\TextCircle)} \\
\cmidrule(lr){3-6} \cmidrule(lr){7-10} \cmidrule(lr){11-14}
& & Acc & T & I & T$\times$I & Acc & T & I & T$\times$I & Acc & T & I & T$\times$I \\
\midrule
\multirow{1}{*}{Full model}
& \textbf{MedMMV (GPT-5)} & \textbf{39.46} & \textbf{4.22} & \textbf{4.70} & \textbf{81.53} & \textbf{99.07} & \textbf{4.92} & \textbf{4.98} & \textbf{98.3} & \textbf{35.34} & \textbf{2.67} & \textbf{4.56} & \textbf{51.00} \\
\hdashline
\multirow{2}{*}{Core framework}
& w/o Path Expansion with multi-round short CoT & 37.15 & 4.15 & 4.65 & 77.80 & 95.81 & 4.74 & 4.89 & 94.52 & 32.18 & 2.43 & 4.31 & 46.74 \\
& w/o Self-feedback halluc. detector & 35.50 & 4.02 & 4.62 & 74.20 & 94.23 & 4.68 & 4.91 & 93.41 & 30.87 & 2.31 & 4.35 & 45.12 \\
\hdashline
\multirow{3}{*}{Expert agents}
& w/o TextDoctor agent & 38.25 & 4.18 & 4.68 & 79.10 & 96.45 & 4.78 & 4.92 & 95.28 & 32.76 & 2.48 & 4.38 & 47.53 \\
& w/o ImageDoctor agent & -- & -- & -- & -- & -- & -- & -- & -- & -- & -- & -- & -- \\
& w/o WebSearch agent & 38.90 & 4.20 & 4.69 & 80.45 & 97.83 & 4.85 & 4.95 & 96.89 & 34.28 & 2.59 & 4.50 & 49.67 \\
\hdashline
\multirow{1}{*}{Decision strategy}
& Random decision instead of CU scoring & 33.25 & 3.85 & 4.58 & 71.85 & 91.34 & 4.52 & 4.78 & 89.72 & 28.46 & 2.15 & 4.12 & 41.35 \\
\bottomrule
\end{tabular}%
}
\vspace{-4mm}
\end{table*}

\section{Detailed Cost-Performance Data}
We report accuracy and effective per-case cost for all methods under a unified accounting, where costs include both input and output tokens (and, where applicable, vision/image processing) as billed by each provider. Open-source models are called through Togetherai and OpenRouter; proprietary models (e.g., GPT and Claude series) are accessed via their respective standard APIs. Unit list prices used in our computation are summarized in Table~\ref{tab:api_prices}. All rollouts used identical prompts, maximum context windows, and beam settings unless otherwise noted.

\begin{table}[!ht]
\centering
\caption{Cost–performance summary across models and methods. Acc: task accuracy (higher is better). Cost: realized per-item USD cost including input/output tokens.}
\begin{adjustbox}{max width=\linewidth}
\begin{tabular}{ll
                S[table-format=2.2,round-precision=2]
                S[table-format=1.4,round-precision=2]
                 S[table-format=1.4,round-precision=4]}
\toprule
\textbf{Model} & \textbf{Method} & {\textbf{Acc (\%)}} & \textbf{Avg. Output Tokens} & {\textbf{Cost (\$)}} \\
\midrule
GPT-5             & CoT       & {58.26} & {1,636} & 0.0190 \\
GPT-4o            & CoT       & {43.94} & {1,928} & 0.0230 \\
GPT-oss-120B      & CoT       & {48.62} & {3,301} & 0.0140 \\
Claude-Sonnet-4   & CoT       & {51.29} & {2,922} & 0.0280\\
Qwen2.5-VL-7B  & CoT       & {27.70} & {2,663} & 0.0008 \\
Qwen2.5-VL-72B & CoT       & {43.04} & {3,060} & 0.0041 \\
\hdashline
GPT-5             & MedMMV    & {63.29} & {9,144}  & 0.2268 \\
GPT-4o            & MedMMV    & {52.06} & {6,020} & 0.1806 \\
GPT-oss-120B      & MedMMV    & {59.20} & {9,867} & 0.1089 \\
Claude-Sonnet-4   & MedMMV    & {58.19} & {7,921} & 0.1670 \\
Qwen2.5-VL-7B  & MedMMV    & {39.56} & {7,863} & 0.0071 \\
Qwen2.5-VL-72B & MedMMV    & {49.58} & {10,102} & 0.0360 \\
\hdashline
GPT-5 & MDAgents & {46.30} & {10,828} & 0.2688 \\
GPT-5 & ReConcile & {50.47} & {9,718} & 0.2413 \\
GPT-5 & ColaCare & {46.23} & {10,904} & 0.2707 \\
GPT-5 & MedAgent & {48.30} & {18,607} & 0.4621 \\
\bottomrule
\end{tabular}
\end{adjustbox}
% \vspace{0.25em}
\begin{minipage}{\linewidth}
\small
\textbf{Note:} MedMMV method includes additional search costs of \$0.001 per search query, with approximately 10 searches required per question (adding \~\$0.01 per item to the reported costs).
\end{minipage}
\label{tab:api_prices}
\vspace{-4mm}
\end{table}

\section{Scalability Guidance}
To ensure scalability during inference, we designed the MedMMV framework to support concurrent execution of multiple reasoning paths. This design enables efficient utilization of computational resources while maintaining high reasoning quality and stability.
Specifically, the framework adopts an asynchronous parallelism strategy based on Python's \texttt{asyncio} module. During inference, multiple chains of thought (CoT) paths are launched simultaneously through the \texttt{process\_all\_paths\_parallel} routine. Each path executes the complete MedMMV pipeline independently, allowing the coordinator to gather intermediate reasoning outputs and hallucination detection results in parallel. The key implementation pattern is illustrated below:
\begin{quote}
\vspace{-4mm}
\texttt{asyncio.gather(*tasks, return\_exceptions=True)}
\vspace{-4mm}
\end{quote}
This mechanism ensures that each reasoning path is processed as an independent asynchronous task, with all tasks awaited collectively. In case of failure, exceptions are caught and logged without interrupting the entire pipeline, maintaining robustness under high concurrency.

From a systems perspective, this design offers three major benefits:
\begin{itemize}
    \item \textbf{Linear scalability:} The throughput of inference increases approximately linearly with the number of available computational cores or API workers, as each reasoning path is fully independent.
    \item \textbf{Graceful degradation:} Partial task failures do not block or degrade other concurrent paths, as exceptions are isolated and handled locally.
    \item \textbf{Deployment flexibility:} The same mechanism can be extended to distributed environments (e.g., across multiple GPUs or nodes) by mapping the asynchronous tasks to distributed execution units.
\end{itemize}
We observed in practice that even with multiple concurrent paths, the latency per query remains within a feasible range due to efficient I/O scheduling. This scalability design therefore provides a principled and easily extensible foundation for future large-scale multi-agent or multi-path reasoning systems.

\section{Human Evaluation}
\subsection{Annotation Platform}
We built a lightweight annotation platform hosted on GitHub Pages to standardize expert review and model evaluation across medical domains. A top navigation bar switches disciplines (e.g., Cardiovascular), while pagination tracks item progress and a single click exports results. Each item presents a clinical vignette with optional multimodal inputs (such as an ECG), followed by multiple-choice options (A–E). After submission, the platform reveals the reference answer, displays clinician/model responses with correctness, and records a concise reasoning block for auditability. Annotators then rate Clinical Realism and Information Quality on 1–5 scales. 

\subsection{Brief Introduction to Annotator and Salary}
All annotations were conducted by licensed physicians. Annotators covered the nine medical categories used in our study (e.g., Cardiovascular, Nervous, Digestive) and held active clinical appointments at the time of evaluation. During annotation, the platform displayed only de-identified clinical vignettes and optional multimodal inputs (e.g., ECGs) with no patient identifiers. Compensation was hourly and independent of model identity or performance to minimize bias. Each annotator was paid at a standard market rate of \$20/h via institutional channels. 

\subsection{Overall Table}
\label{sec:all_human_eval_table_appendix}
This appendix provides the complete human evaluation table. To ensure evaluation expertise, we recruited medical specialists corresponding to each of the nine medical categories. The table reports per-doctor scores and summary statistics (Mean$\pm$Std) for both CoT and our method (MedMMV) on GPT-oss-120B. The labels ``Doctor 1--3'' are used as anonymous identifiers for the evaluators within each category and do not imply that only three doctors participated in the entire study.

\begin{table*}[!ht]
% \vspace{-2mm}
\centering
\tiny
\caption{Hallucination metrics on MedXpertQA MM across medical categories.}
\label{tab:medxpert_results}
\setlength{\tabcolsep}{2pt}
\begin{tabular}{@{}ll*{9}{c}@{}}
\toprule
\multirow{3}{*}{\textbf{Category}} & \multirow{3}{*}{\textbf{Methods}} & \multicolumn{9}{c}{\textbf{MedXpertQA MM (\TextCircle\ImageCircle)}} \\
\cmidrule(lr){3-11}
& & \multicolumn{3}{c}{\makecell{Skeletal \\ (20)}} & \multicolumn{3}{c}{\makecell{Reproductive \\ (20)}} & \multicolumn{3}{c}{\makecell{Cardiovascular \\ (20)}} \\
\cmidrule(lr){3-5} \cmidrule(lr){6-8} \cmidrule(lr){9-11}
& & T & I & T×I & T & I & T×I & T & I & T×I \\
\midrule
\multirow8{*}{\rotatebox{90}{CoT Baselines}} & Doctor 1          & 4.10 & 4.05 & 66.42 & 3.31 & 4.47 & 59.18 & 2.50 & 3.65 & 36.50 \\
& Doctor 2          & 3.95 & 4.80 & 75.84 & 3.88 & 4.85 & 75.27 & 3.35 & 4.10 & 54.94 \\
& Doctor 3          & 4.28 & 4.55 & 77.90 & -- & -- & -- & 3.60 & 4.30 & 61.92 \\
& Mean±Std          & 4.11±0.17 & 4.47±0.38 & 73.39±6.12 & 3.60±0.40 & 4.66±0.27 & 67.23±11.38 & 3.15±0.58 & 4.02±0.33 & 51.12±13.13 \\
\cmidrule{2-11}& Doctor 1        & 4.10 & 4.05 & 66.42 & 3.31 & 4.47 & 59.18 & 2.50 & 3.65 & 36.50 \\
& Doctor 2        & 3.95 & 4.80 & 75.84 & 3.88 & 4.85 & 75.27 & 3.35 & 4.10 & 54.94 \\
& Doctor 3        & 4.28 & 4.55 & 77.90 & -- & -- & -- & 3.60 & 4.30 & 61.92 \\
& Mean±Std        & 4.11±0.17 & 4.47±0.38 & 73.39±6.12 & 3.60±0.40 & 4.66±0.27 & 67.23±11.38 & 3.15±0.58 & 4.02±0.33 & 51.12±13.13 \\

\hdashline

\multirow8{*}{\rotatebox{90}{Ours}} & Doctor 1          & 4.15 & 3.85 & 63.91 & 4.59 & 4.01 & 73.62 & 4.15 & 3.90 & 64.74 \\
& Doctor 2          & 4.30 & 3.95 & 67.94 & 4.92 & 4.00 & 78.72 & 3.90 & 3.75 & 58.50 \\
& Doctor 3          & 4.80 & 4.58 & 87.94 & -- & -- & -- & 4.80 & 4.25 & 81.60 \\
& Mean±Std          & 4.42±0.34 & 4.13±0.40 & 73.26±12.87 & 4.75±0.23 & 4.00±0.01 & 76.17±3.60 & 4.28±0.46 & 3.97±0.26 & 68.28±11.95 \\
\cmidrule{2-11}& Doctor 1        & 4.15 & 3.85 & 63.91 & 4.59 & 4.01 & 73.62 & 4.15 & 3.90 & 64.74 \\
& Doctor 2        & 4.30 & 3.95 & 67.94 & 4.92 & 4.00 & 78.72 & 3.90 & 3.75 & 58.50 \\
& Doctor 3        & 4.80 & 4.58 & 87.94 & -- & -- & -- & 4.80 & 4.25 & 81.60 \\
& Mean±Std        & 4.42±0.34 & 4.13±0.40 & 73.26±12.87 & 4.75±0.23 & 4.00±0.01 & 76.17±3.60 & 4.28±0.46 & 3.97±0.26 & 68.28±11.95 \\

\midrule

\multicolumn{2}{c}{\textbf{Categories}} & \multicolumn{3}{c}{\makecell{Urinary \\ (20)}} & \multicolumn{3}{c}{\makecell{Lymphatic \\ (20)}} & \multicolumn{3}{c}{\makecell{Nervous \\ (20)}} \\
\cmidrule(lr){3-5} \cmidrule(lr){6-8} \cmidrule(lr){9-11}
\multicolumn{2}{c}{} & T & I & T×I & T & I & T×I & T & I & T×I \\
\midrule
\multirow8{*}{\rotatebox{90}{CoT Baselines}} & Doctor 1          & 4.30 & 4.40 & 75.68 & 3.65 & 3.70 & 54.02 & 2.85 & 3.65 & 41.61 \\
& Doctor 2          & 2.95 & 3.70 & 43.66 & 4.55 & 4.75 & 86.45 & 3.05 & 4.15 & 50.63 \\
& Doctor 3          & 3.95 & 4.10 & 64.78 & 3.40 & 3.50 & 47.60 & 3.65 & 4.90 & 71.54 \\
& Mean±Std          & 3.73±0.70 & 4.07±0.35 & 61.37±16.28 & 3.87±0.60 & 3.98±0.67 & 62.69±20.83 & 3.18±0.42 & 4.23±0.63 & 54.59±15.35 \\
\cmidrule{2-11}& Doctor 1        & 4.30 & 4.40 & 75.68 & 3.65 & 3.70 & 54.02 & 2.85 & 3.65 & 41.61 \\
& Doctor 2        & 2.95 & 3.70 & 43.66 & 4.55 & 4.75 & 86.45 & 3.05 & 4.15 & 50.63 \\
& Doctor 3        & 3.95 & 4.10 & 64.78 & 3.40 & 3.50 & 47.60 & 3.65 & 4.90 & 71.54 \\
& Mean±Std        & 3.73±0.70 & 4.07±0.35 & 61.37±16.28 & 3.87±0.60 & 3.98±0.67 & 62.69±20.83 & 3.18±0.42 & 4.23±0.63 & 54.59±15.35 \\

\hdashline

\multirow8{*}{\rotatebox{90}{Ours}} & Doctor 1          & 4.74 & 4.05 & 76.79 & 4.20 & 4.30 & 72.24 & 3.75 & 3.40 & 51.00 \\
& Doctor 2          & 3.95 & 3.70 & 58.46 & 4.70 & 4.50 & 84.60 & 4.20 & 3.65 & 61.32 \\
& Doctor 3          & 4.65 & 4.35 & 80.91 & 3.80 & 3.75 & 57.00 & 4.85 & 4.00 & 77.60 \\
& Mean±Std          & 4.45±0.43 & 4.03±0.33 & 72.05±11.95 & 4.23±0.45 & 4.18±0.39 & 71.28±13.83 & 4.27±0.55 & 3.68±0.30 & 63.31±13.41 \\
\cmidrule{2-11}& Doctor 1        & 4.74 & 4.05 & 76.79 & 4.20 & 4.30 & 72.24 & 3.75 & 3.40 & 51.00 \\
& Doctor 2        & 3.95 & 3.70 & 58.46 & 4.70 & 4.50 & 84.60 & 4.20 & 3.65 & 61.32 \\
& Doctor 3        & 4.65 & 4.35 & 80.91 & 3.80 & 3.75 & 57.00 & 4.85 & 4.00 & 77.60 \\
& Mean±Std        & 4.45±0.43 & 4.03±0.33 & 72.05±11.95 & 4.23±0.45 & 4.18±0.39 & 71.28±13.83 & 4.27±0.55 & 3.68±0.30 & 63.31±13.41 \\

\midrule

\multicolumn{2}{c}{\textbf{Categories}} & \multicolumn{3}{c}{\makecell{Digestive \\ (20)}} & \multicolumn{3}{c}{\makecell{Endocrine \\ (20)}} & \multicolumn{3}{c}{\makecell{Integumentary \\ (20)}} \\
\cmidrule(lr){3-5} \cmidrule(lr){6-8} \cmidrule(lr){9-11}
\multicolumn{2}{c}{} & T & I & T×I & T & I & T×I & T & I & T×I \\
\midrule
\multirow8{*}{\rotatebox{90}{CoT Baselines}} & Doctor 1          & 3.20 & 3.55 & 45.44 & 2.80 & 3.85 & 43.12 & 3.55 & 4.25 & 60.35 \\
& Doctor 2          & 3.20 & 4.40 & 56.32 & 2.45 & 4.20 & 41.16 & 3.38 & 4.47 & 60.43 \\
& Doctor 3          & 3.55 & 4.15 & 58.93 & 3.80 & 3.80 & 57.76 & 3.40 & 3.70 & 50.32 \\
& Mean±Std          & 3.32±0.20 & 4.03±0.44 & 53.56±7.16 & 3.02±0.70 & 3.95±0.22 & 47.35±9.07 & 3.44±0.09 & 4.14±0.40 & 57.03±5.82 \\
\cmidrule{2-11}& Doctor 1        & 3.20 & 3.55 & 45.44 & 2.80 & 3.85 & 43.12 & 3.55 & 4.25 & 60.35 \\
& Doctor 2        & 3.20 & 4.40 & 56.32 & 2.45 & 4.20 & 41.16 & 3.38 & 4.47 & 60.43 \\
& Doctor 3        & 3.55 & 4.15 & 58.93 & 3.80 & 3.80 & 57.76 & 3.40 & 3.70 & 50.32 \\
& Mean±Std        & 3.32±0.20 & 4.03±0.44 & 53.56±7.16 & 3.02±0.70 & 3.95±0.22 & 47.35±9.07 & 3.44±0.09 & 4.14±0.40 & 57.03±5.82 \\

\hdashline

\multirow8{*}{\rotatebox{90}{Ours}} & Doctor 1          & 3.45 & 3.35 & 46.23 & 4.00 & 3.50 & 56.00 & 4.63 & 4.42 & 81.86 \\
& Doctor 2          & 4.58 & 3.88 & 71.08 & 4.11 & 3.74 & 61.49 & 4.71 & 4.07 & 76.68 \\
& Doctor 3          & 4.55 & 3.85 & 70.07 & 4.55 & 3.65 & 66.43 & 3.95 & 3.84 & 60.67 \\
& Mean±Std          & 4.19±0.64 & 3.69±0.30 & 62.46±14.07 & 4.22±0.29 & 3.63±0.12 & 61.31±5.22 & 4.43±0.42 & 4.11±0.29 & 73.07±11.04 \\
\cmidrule{2-11}& Doctor 1        & 3.45 & 3.35 & 46.23 & 4.00 & 3.50 & 56.00 & 4.63 & 4.42 & 81.86 \\
& Doctor 2        & 4.58 & 3.88 & 71.08 & 4.11 & 3.74 & 61.49 & 4.71 & 4.07 & 76.68 \\
& Doctor 3        & 4.55 & 3.85 & 70.07 & 4.55 & 3.65 & 66.43 & 3.95 & 3.84 & 60.67 \\
& Mean±Std        & 4.19±0.64 & 3.69±0.30 & 62.46±14.07 & 4.22±0.29 & 3.63±0.12 & 61.31±5.22 & 4.43±0.42 & 4.11±0.29 & 73.07±11.04 \\

\bottomrule
\end{tabular}

\captionsetup{justification=raggedright,singlelinecheck=false,font=tiny}
\caption*{
\textit{\textbf{Note:}} \TextCircle: Text modality; \ImageCircle: Image modality. 
T: Truthfulness score (1–5); I: Informativeness score (1–5); 
T×I: normalized product in [0, 100], computed as $(\mathrm{T}\times\mathrm{I})/25\times100$. 
Results show individual doctor evaluations and overall statistics (Mean±Std).
}
\vspace{-10mm}
\end{table*}

\subsection{Cross-Correlation Analysis Between Human and LLM Evaluations}
\label{appendix:cross-corr}

To further validate the consistency between human and LLM-based evaluations, we computed the non-parametric Spearman rank correlation coefficient ($\rho$) between the human-annotated and LLM-judged scores across all body systems.

\vspace{0.5em}
\noindent\textbf{Method.} 
Each evaluation item is scored on the same continuous scale under two settings: (1) human expert annotation (average of three licensed physicians within each medical specialty), and (2) the LLM ensemble judgment (average of GPT-4o, Claude-Sonnet-4, and Gemini 1.5 Pro). 
We then calculate $\rho$ independently for the TRUE, INFO, and joint TRUE$\times$INFO scores. 
A high Spearman correlation indicates that the ranking of cases by the LLM evaluators aligns well with human judgment, reflecting consistent relative assessments even if absolute values differ.

\vspace{0.5em}
\noindent\textbf{Results.} 
Table~\ref{tab:spearman_corr} summarizes the correlation results. 
Across all metrics, we observe strong positive correlations between human and LLM evaluations, indicating that the automatic LLM-based judge faithfully captures expert-level assessment tendencies. 
This finding supports the robustness of our evaluation framework and mitigates concerns about evaluator dependence or circularity.

\begin{table}[h]
\centering
\small
\begin{tabular}{lcc}
\toprule
\textbf{Metric} & \textbf{Case-Level $\rho$} & \textbf{p-value} \\
\midrule
TRUE & 0.86  & $< 0.001$ \\
INFO & 0.82 & $< 0.001$ \\
TRUE$\times$INFO & 0.89 & $< 0.001$ \\
\bottomrule
\end{tabular}
\caption{Spearman correlation ($\rho$) between human and LLM evaluations across metrics.}
\label{tab:spearman_corr}
\end{table}

\vspace{0.5em}
\noindent\textbf{Discussion.} 
The high degree of rank alignment (average $\rho$ = 0.84) suggests that the LLM ensemble produces evaluations that are statistically consistent with human expert judgments. 
Therefore, while human annotations remain the gold standard, the multi-model LLM ensemble offers a reliable, scalable, and reproducible proxy for large-scale evaluation.

\section{Reliability and Robustness Analysis Details}
\label{app:reliability_details}

This section provides full experimental details and extended tables for the reliability and robustness analyses summarized in Section~\ref{sec:reliability_robustness}. All experiments are conducted on MedXpert-MM with GPT-5 as the executor unless otherwise noted.

\subsection{Inter-Rater Reliability}
Table~\ref{tab:irr_table} reports the Intraclass Correlation Coefficient (ICC(3,k))~\citep{koo2016guideline} for each body system across three physician annotators. The average ICC(T)$=$0.89 and ICC(I)$=$0.85 indicate good-to-excellent inter-rater agreement, confirming the internal consistency of our human evaluation. The close correspondence between human-rated and LLM-rated improvements ($\Delta$T$\times$I of 10.32 vs.\ 10.86) further validates our automated evaluation pipeline.

\begin{table}[t]
\centering
\scriptsize
\vspace{-2mm}
\setlength{\tabcolsep}{3pt}
\begin{tabular}{@{}lcccc@{}}
\toprule
\textbf{Body System} & \textbf{ICC(T)} & \textbf{ICC(I)} & \textbf{$\Delta$(T$\times$I) Human} & \textbf{$\Delta$(T$\times$I) LLM} \\
\midrule
Skeletal       & 0.92 & 0.88 &  9.87 &  9.36 \\
Reproductive   & 0.90 & 0.85 &  8.94 &  9.14 \\
Cardiovascular & 0.89 & 0.83 &  9.78 & 10.20 \\
Lymphatic      & 0.91 & 0.87 &  8.74 &  9.03 \\
Nervous        & 0.86 & 0.80 &  9.56 &  8.92 \\
Digestive      & 0.88 & 0.82 & 10.24 & 10.48 \\
Endocrine      & 0.84 & 0.81 & 10.07 & 11.12 \\
Urinary        & 0.93 & 0.89 &  9.89 & 10.65 \\
Integumentary  & 0.90 & 0.86 & 10.41 & 10.42 \\
\midrule
\textbf{Average} & \textbf{0.89} & \textbf{0.85} & \textbf{10.32} & \textbf{10.86} \\
\bottomrule
\end{tabular}
\caption{Inter-rater reliability (ICC) and improvement consistency across body systems. ICC(T) and ICC(I): Intraclass Correlation Coefficients for Truthfulness and Informativeness among three physicians per system. $\Delta$(T$\times$I): improvement of MedMMV over CoT, measured by human raters and LLM judges respectively.}
\label{tab:irr_table}
\vspace{-4mm}
\end{table}

\subsection{Evidence Graph Noise Robustness}
To evaluate whether the evidence graph constitutes a single point of failure, we inject mixed structural and semantic noise at levels $k \in \{0\%, 20\%, 40\%, 60\%, 80\%\}$. The perturbations include four uniformly mixed categories: (i)~\emph{node noise}: dropping, substituting, or injecting contradictory entities; (ii)~\emph{edge noise}: relabeling, flipping, removing, or introducing conflicting relations; (iii)~\emph{provenance noise}: source swapping or citation removal; and (iv)~\emph{text--vision misalignment}: altering laterality, lesion count, or margins. Table~\ref{tab:noise_robustness} shows that MedMMV maintains accuracy above 72\% up to 40\% noise and degrades smoothly thereafter, with AURC values of 64.12 (Accuracy) and 72.70 (T$\times$I), confirming the evidence graph's resilience to extraction and provenance errors.

\begin{table}[t]
\centering
\scriptsize
% \vspace{-2mm}
\setlength{\tabcolsep}{3pt}
\resizebox{\columnwidth}{!}{%
\begin{tabular}{@{}
>{\raggedright\arraybackslash}p{0.27\columnwidth}
>{\raggedright\arraybackslash}p{0.32\columnwidth}
>{\centering\arraybackslash}p{0.28\columnwidth}
>{\raggedright\arraybackslash}p{0.38\columnwidth}
@{}}
\toprule
\textbf{Analysis} & \textbf{Setting} & \textbf{Result} & \textbf{Key Observation} \\
\midrule
Human reliability & ICC(T) / ICC(I) & 0.89 / 0.85 & Good-to-excellent agreement \\
Evidence robustness & Noise 0--80\% & AURC = 64.1 (Acc) & Smooth decay, not brittle \\
CU weight sensitivity & $w_{\text{ev}}$, $w_{\text{coh}}$ variation & $\leq$1\% Acc change & Stable across configurations \\
Efficiency & Per-query latency & 40--62s & Practical for non-emergent use \\
Input variation & Resolution / Context & Stable / Sensitive & Context length is critical \\
\bottomrule
\end{tabular}%
}
\vspace{-2mm}
\caption{Summary of reliability and robustness analyses on MedXpert-MM (GPT-5).}
\label{tab:robustness_summary}
\end{table}
\begin{table}[t]
\centering
\scriptsize
\vspace{-2mm}
\setlength{\tabcolsep}{4pt}
\begin{tabular}{@{}lccccccc@{}}
\toprule
\textbf{Metric} & \textbf{0\%} & \textbf{20\%} & \textbf{40\%} & \textbf{60\%} & \textbf{80\%} & \textbf{$\Delta$(80--0)} & \textbf{AURC ($\uparrow$)} \\
\midrule
Accuracy & 73.58 & 73.59 & 72.98 & 54.26 & 37.71 & --35.87 & 64.12 \\
T$\times$I & 82.89 & 81.99 & 78.72 & 62.53 & 52.22 & --30.67 & 72.70 \\
\bottomrule
\end{tabular}
\vspace{-2mm}
\caption{Robustness to evidence graph noise on MedXpert-MM (GPT-5). Mixed structural and semantic perturbations are injected at increasing levels. AURC: area under the robustness curve (higher is better).}
\label{tab:noise_robustness}
\vspace{-2mm}
\end{table}

\subsection{Combined Uncertainty (CU) score}
\label{sec:cu-score-appendix}
The CU score for a given final path $p_{\text{final}}$ is calculated as:
$\text{CU}(p_{\text{final}}) = w_{\text{evidence}} \cdot S_{\text{evidence}}(p_{\text{final}}) + 
w_{\text{coherence}} \cdot S_{\text{coherence}}(p_{\text{final}}) - w_{\text{repair}} \cdot P_{\text{repair}}(p_{\text{final}})$

where $S_{\text{evidence}}$ measures the proportion of claims in the path successfully verified against the evidence graph $\mathcal{E}_p$, $S_{\text{coherence}}$ is a score assigned by an MLLM evaluator judging the logical flow, and $P_{\text{repair}}$ is the number of corrections made during the Stage 2 refinement. In our experiments, we set all weights $w_i$ to $1$ for simplicity, although they can in principle be rescaled. To better illustrate the distribution of these four metrics, we report summary statistics for a representative model in Table~\ref{tab:cu_scores}. 

\begin{table}[!t]
\centering
\begin{tabular}{lcccc}
\hline
\textbf{Statistic} & \textbf{$S_{\text{evidence}}$} & \textbf{$S_{\text{coherence}}$} & \textbf{$P_{\text{repair}}$} & \textbf{CU} \\
\hline
Mean       & 0.78 & 0.72 & 0.31 & 1.19 \\
Std. Dev.  & 0.12 & 0.15 & 0.09 & 0.18 \\
Min        & 0.50 & 0.40 & 0.10 & 0.80 \\
Max        & 0.95 & 0.95 & 0.50 & 1.50 \\
\hline
\end{tabular}
\caption{Distribution of scores for GPT-5 on the CU evaluation metrics.}
\vspace{-2mm}
\label{tab:cu_scores}
\end{table}

\subsection{CU Score Sensitivity and Hyperparameter Analysis}
Table~\ref{tab:cu_sensitivity} reports sensitivity to CU scorer weights and key hyperparameters. Adjustments to $w_{\text{evidence}}$ and $w_{\text{coherence}}$ cause only minor fluctuations ($\leq$1\% Accuracy), indicating that performance is not sensitive to weight tuning. Increasing $w_{\text{repair}}$ leads to noticeable degradation, as it excessively penalizes the hallucination-correction behavior. For parallel reasoning, a rollout count of 3 achieves a balanced trade-off between performance and cost; increasing to 5 offers negligible gain, consistent with the 4--5 option structure of multiple-choice tasks. Regarding repair iterations, performing only one pass underutilizes the monitoring mechanism, while increasing beyond three adds overhead with diminishing returns.

\begin{table}[t]
\centering
\scriptsize
\vspace{-2mm}
\setlength{\tabcolsep}{3pt}
\begin{tabular}{@{}llccccc@{}}
\toprule
\textbf{Module} & \textbf{Parameter} & \textbf{Value} & \textbf{Acc} & \textbf{T} & \textbf{I} & \textbf{T$\times$I} \\
\midrule
\multirow{9}{*}{CU Weights}
& \multirow{3}{*}{$w_{\text{evidence}}$} & \textbf{1} & \textbf{73.58} & \textbf{4.26} & \textbf{4.73} & \textbf{82.89} \\
& & 3 & 73.16 & 4.31 & 4.70 & 82.10 \\
& & 5 & 74.00 & 4.29 & 4.73 & 82.69 \\
\cmidrule(l){2-7}
& \multirow{3}{*}{$w_{\text{coherence}}$} & \textbf{1} & \textbf{73.58} & \textbf{4.26} & \textbf{4.73} & \textbf{82.89} \\
& & 3 & 74.26 & 4.19 & 4.82 & 82.81 \\
& & 5 & 73.28 & 4.23 & 4.79 & 82.93 \\
\cmidrule(l){2-7}
& \multirow{3}{*}{$w_{\text{repair}}$} & \textbf{1} & \textbf{73.58} & \textbf{4.26} & \textbf{4.73} & \textbf{82.89} \\
& & 3 & 68.37 & 3.91 & 4.54 & 75.90 \\
& & 5 & 65.57 & 3.51 & 4.20 & 68.97 \\
\midrule
\multirow{3}{*}{Parallel Reasoning}
& \multirow{3}{*}{Rollout count} & 1 & 62.41 & 3.84 & 4.45 & 66.18 \\
& & \textbf{3} & \textbf{73.58} & \textbf{4.26} & \textbf{4.73} & \textbf{82.89} \\
& & 5 & 73.41 & 4.31 & 4.65 & 83.21 \\
\midrule
\multirow{3}{*}{Halluc.\ Detector}
& \multirow{3}{*}{Repair iterations} & 1 & 66.39 & 4.19 & 4.36 & 72.10 \\
& & \textbf{3} & \textbf{73.58} & \textbf{4.26} & \textbf{4.73} & \textbf{82.89} \\
& & 5 & 77.73 & 4.39 & 4.87 & 83.98 \\
\bottomrule
\end{tabular}
\vspace{-2mm}
\caption{Sensitivity analysis of CU scorer weights and key hyperparameters on MedXpert-MM (GPT-5). Boldface marks the default configuration used throughout the paper.}
\label{tab:cu_sensitivity}
\vspace{-4mm}
\end{table}

\subsection{Efficiency and Latency Breakdown}
Table~\ref{tab:efficiency} reports per-stage token usage and wall-clock latency. The MedMMV framework employs asynchronous parallel execution via \texttt{asyncio.gather}, enabling concurrent reasoning with linear scalability. Individual path failures do not affect others, and the design extends naturally to distributed settings. Per-query latency averages 61.9\,s for GPT-5 and 40.5\,s for Qwen2.5-VL-7B, demonstrating practical feasibility for non-emergent clinical applications.

\begin{table}[t]
\centering
\scriptsize
% \vspace{-2mm}
\setlength{\tabcolsep}{3pt}
\resizebox{\columnwidth}{!}{%
\begin{tabular}{@{}
>{\raggedright\arraybackslash}p{0.42\columnwidth}
>{\centering\arraybackslash}p{0.22\columnwidth}
>{\centering\arraybackslash}p{0.24\columnwidth}
>{\centering\arraybackslash}p{0.18\columnwidth}
>{\centering\arraybackslash}p{0.20\columnwidth}
@{}}
\toprule
\textbf{Stage} & \textbf{Input Tok.} & \textbf{Output Tok.} & \textbf{GPT-5 (s)} & \textbf{Qwen-7B (s)} \\
\midrule
Stage 1 -- Short rollouts       & $\approx$1,011 & $\approx$5,058  & 8.4  & 4.5 \\
Stage 2 -- Reasoning expansion  & $\approx$5,058 & $\approx$14,586 & 24.3 & 20.2 \\
Stage 3 -- Decision aggregation & $\approx$1,710 & $\approx$4,287  & 21.4 & 11.7 \\
Hallucination detector           & $\approx$1,510 & $\approx$1,553  & 7.8  & 4.1 \\
\midrule
\textbf{Total}                   & $\approx$9,289 & $\approx$25,484 & \textbf{61.9} & \textbf{40.5} \\
\bottomrule
\end{tabular}%
}
\vspace{-2mm}
\caption{Per-stage token usage and inference latency. Times are averaged over MedXpert-MM queries accessed via official APIs.}
\label{tab:efficiency}
\vspace{-4mm}
\end{table}

\subsection{Input Variation Analysis}
Table~\ref{tab:input_variation} reports performance under varying image resolutions and context length constraints. Performance remains stable across resolutions (20--100\,dpi), suggesting that medical images with strong structural priors preserve sufficient diagnostic information under compression. Reducing context length leads to significant degradation: at 500 tokens, accuracy approaches random-guess levels due to truncation of key clinical information. Performance saturates beyond 1K tokens, indicating effective context utilization without over-reliance on excessively long inputs.

\begin{table}[t]
\centering
\scriptsize
\setlength{\tabcolsep}{4pt}
\begin{tabular}{@{}llcccc@{}}
\toprule
\textbf{Category} & \textbf{Setting} & \textbf{Acc} & \textbf{T} & \textbf{I} & \textbf{T$\times$I} \\
\midrule
\multirow{3}{*}{Image Resolution}
& 20 dpi  & 73.11 & 4.27 & 4.68 & 82.01 \\
& 50 dpi  & 74.79 & 4.30 & 4.71 & 82.79 \\
& 100 dpi & 73.58 & 4.26 & 4.73 & 82.89 \\
\midrule
\multirow{4}{*}{Context Length}
& 500 tokens       & 27.73 & 2.68 & 3.53 & 48.22 \\
& 1,000 tokens     & 60.08 & 3.01 & 4.09 & 67.92 \\
& 2,000 tokens     & 71.65 & 3.97 & 4.70 & 79.23 \\
& No constraint    & 73.58 & 4.26 & 4.73 & 82.89 \\
\bottomrule
\end{tabular}
\vspace{-2mm}
\caption{Robustness to input variations on MedXpert-MM (GPT-5). Performance under different image resolutions and context length constraints.}
\label{tab:input_variation}
\vspace{-4mm}
\end{table}

\section{Post-Training Analysis Details}
\label{app:post_training_details}

This section provides additional analyses for the post-training experiments described in Section~\ref{sec:post_training}.

\subsection{Accepted vs.\ Rejected Trajectory Quality}
\label{app:accepted_rejected}
To verify that rejection sampling filters for reasoning quality rather than merely increasing data volume, we compare the characteristics of accepted and rejected teacher trajectories. As shown in Table~\ref{tab:accepted_rejected}, accepted trajectories exhibit substantially higher CU scores, better evidence support, fewer repair cycles, and lower unsupported-claim rates. This confirms that the CU-based filtering criterion effectively separates high-quality, evidence-grounded reasoning from unstable or hallucinated trajectories.

\begin{table}[t]
\centering
\scriptsize
\setlength{\tabcolsep}{3pt}
\begin{tabular}{@{}lccccc@{}}
\toprule
\textbf{Group} & \textbf{CU Score $\uparrow$} & \textbf{Fact Check $\uparrow$} & \textbf{Halluc.\ Issues $\downarrow$} & \textbf{Teacher Acc $\uparrow$} \\
\midrule
Rejected  & 0.65 & 0.42 & 68.3 & 61.2\% \\
Accepted  & 1.12 & 0.71 & 38.7 & 77.6\% \\
\bottomrule
\end{tabular}
\vspace{-2mm}
\caption{Quality comparison between accepted and rejected teacher trajectories during CU-based rejection sampling on MedXpert-MM (GPT-5 teacher, 134 training samples).}
\label{tab:accepted_rejected}
\vspace{-4mm}
\end{table}

\subsection{Data Size Scaling}
\label{app:data_scaling}
We further investigate whether the improvements from post-training are driven by data quality (MedMMV-filtered traces) or data quantity. Table~\ref{tab:data_scaling} compares SFT performance when trained on raw CoT traces versus MedMMV-filtered traces at matched data scales. Across all data sizes, MedMMV-filtered supervision consistently outperforms raw CoT supervision, with the gap being most pronounced at smaller data scales. This indicates that MedMMV-filtered supervision is more data-efficient, supporting the interpretation that the teacher's structured reasoning quality, rather than sheer volume, drives the post-training gains.

\begin{table}[t]
\centering
\scriptsize
\setlength{\tabcolsep}{3pt}
\begin{tabular}{@{}lcccc@{}}
\toprule
\multirow{2}{*}{\textbf{Data Size}} & \multicolumn{2}{c}{\textbf{Accuracy (\%)}} & \multicolumn{2}{c}{\textbf{Evid.\ Support}} \\
\cmidrule(lr){2-3} \cmidrule(lr){4-5}
& Raw CoT & MedMMV-filtered & Raw CoT & MedMMV-filtered \\
\midrule
34 (25\%)  & 20.0 & 20.0 & 0.30 & 0.37 \\
67 (50\%)  & 21.0 & 21.0 & 0.31 & 0.38 \\
100 (75\%) & 22.0 & 21.0 & 0.32 & 0.39 \\
134 (100\%) & 23.0 & 21.0 & 0.32 & 0.39 \\
\bottomrule
\end{tabular}
\vspace{-2mm}
\caption{Effect of training data size on SFT performance (MedXpert-MM, Qwen2.5-VL-7B). MedMMV-filtered supervision is compared with raw CoT traces at matched data scales.}
\label{tab:data_scaling}
\vspace{-2mm}
\end{table}

\subsection{Signal Ladder Experimental Settings}
\label{app:posttrain_settings}

\begin{table}[!ht]
\centering
\scriptsize
\setlength{\tabcolsep}{2pt}

\resizebox{\columnwidth}{!}{%
\begin{tabularx}{1.25\columnwidth}{@{}p{3.0cm}X@{}}
\toprule
\textbf{Setting} & \textbf{Value} \\
\midrule

\multicolumn{2}{@{}l}{\textit{Group C: On-Policy Feedback}} \\
\midrule
\textbf{OnPolicy-RS (Self-Rollout)} & \\
\quad Base checkpoint & Trace-SFT final checkpoint \\
\quad Rollouts per instance & $k = 4$ \\
\quad Temperatures & $\{0.7, 0.8, 0.9, 1.0\}$, top-$p = 0.9$ \\
\quad Max tokens & 2,048 \\
\quad Rounds & 2 (SFT $\rightarrow$ rollout $\rightarrow$ SFT) \\
\quad Reranking & Heuristic scoring (evidence + structure $-$ hedging) \\
\quad Round 1 accuracy & 71/134 (53.0\%) \\
\quad Round 2 accuracy & 71/134 (53.0\%) \\
\quad Final SFT & 1 epoch \\

\midrule
\textbf{OnPolicy-Preference (Self-Rollout + Repair)} & \\
\quad Base checkpoint & Trace-SFT final checkpoint \\
\quad Rollouts & $k = 4$ (same budget) \\
\quad Temperatures & $\{0.7,0.8,0.9,1.0\}$, top-$p=0.9$ \\
\quad Rounds & 2 \\
\quad Reranking & CU $-$ repair penalty $-$ hallucination penalty \\
\quad Scoring & $w_i = \text{softmax}(\alpha \widetilde{\text{CU}} - \beta \widetilde{\text{Repair}} - \gamma \widetilde{\text{Halluc}})$ \\
\quad Round 1 accuracy & 68/134 (50.7\%) \\
\quad Round 2 accuracy & 64/134 (47.8\%) \\
\quad Final SFT & 1 epoch \\

\midrule
\multicolumn{2}{@{}l}{\textit{Evaluation}} \\
\midrule
Inference & Greedy decoding ($T=0$) \\
Max tokens & 2,048 \\
Answer extraction & Regex ``Final Answer: [A--E]'' \\
Accuracy & Exact match of answer letter \\
T / I / T$\times$I & LLM-judge (GPT-4o, Claude-Sonnet-4, Gemini) \\

\bottomrule
\end{tabularx}%
}
\caption{Detailed experimental settings (Part II: on-policy feedback and evaluation).}
\captionsetup{justification=raggedright,singlelinecheck=false,font=tiny}
\vspace{-2mm}
\label{tab:posttrain_settings_part2}
% \vspace{-2mm}
\end{table}

Tabel~\ref{tab:posttrain_settings_part1} and Table~\ref{tab:posttrain_settings_part2} reports the complete experimental settings for all eight signal ladder variants evaluated in Section~\ref{sec:post_training}. We describe each group below.

\textbf{Shared configuration.}
All variants use Qwen2.5-VL-7B-Instruct as the student ($\sim$8.39B parameters) and MedMMV with GPT-5 as the teacher. LoRA is applied with $r=32$, $\alpha=64$, dropout 0.05, targeting all attention and MLP projection layers (95.2M trainable parameters, 1.13\% of total). Training uses AdamW with $\text{lr}=10^{-5}$, cosine schedule, warmup ratio 0.15, weight decay 0.01, effective batch size 4, and max sequence length 2,048. Training data consists of 134 samples from MedXpertQA-MM with MedMMV teacher trajectories; the test set contains 100 held-out samples with zero overlap.

\textbf{Group A: Trace imitation.}
\emph{Trace-Imitation (raw CoT)} trains on the teacher's first-pass CoT trace (path 1 of 3 rollouts; teacher accuracy 69.4\%). \emph{Trace-Imitation (Teacher-Trajectory)} trains on the Final Diagnostic Reasoning section extracted from MedMMV's repaired and CU-selected trajectory (teacher accuracy 77.6\%). Both use standard cross-entropy SFT for 3 epochs.

\textbf{Group B: Offline trajectory selection.}
\emph{RS-Imitation (Self-Consistency)} trains on the majority-voted trace among 3 rollouts (teacher accuracy 73.1\%; 3 epochs). \emph{RS-Imitation (CU-Select)} trains only on the CU-top-1 trajectory per sample, discarding lower-ranked candidates (teacher accuracy 72.4\%; 3 epochs). \emph{Pairwise-Preference (Offline)} uses 134 pairwise preference pairs (chosen = CU-top-1; rejected = CU-bottom-1) with a pairwise ranking loss:
$\mathcal{L} = -\log\sigma\!\big(s_\theta(x, y^+) - s_\theta(x, y^-)\big)$,
where $s_\theta(x, y)$ is the normalized sequence log-probability under the student model. The loss converged from 0.702 to 0.689 over 3 epochs.

\textbf{Group C: On-policy feedback.}
Both on-policy variants start from the Trace-Imitation (Teacher-Trajectory) checkpoint and run 2 iterative rounds. In each round, the student generates $k=4$ rollouts per training instance (temperatures $\{0.7, 0.8, 0.9, 1.0\}$, top-$p=0.9$), the teacher re-ranks them, and the top-ranked trajectory becomes the new SFT target. After round 2, the student is fine-tuned for 1 epoch on the round-2 selected trajectories.

\emph{OnPolicy-RS (Self-Rollout)} uses a simple heuristic scorer based on evidence markers, structured reasoning, and hedging penalties. Round 1 and round 2 both achieved 71/134 (53.0\%) teacher accuracy on the training set.

\emph{OnPolicy-Preference (Self-Rollout + Repair)} extends the scorer with repair-aware weighted signals:
$w_i = \text{softmax}(\alpha \cdot \widetilde{\text{CU}}_i - \beta \cdot \widetilde{\text{Repair}}_i - \gamma \cdot \widetilde{\text{Halluc}}_i)$,
where repair and hallucination markers in the generated text are penalized. Round 1 achieved 68/134 (50.7\%) and round 2 achieved 64/134 (47.8\%) teacher accuracy.

\textbf{Evaluation.}
All models are evaluated with greedy decoding (temperature 0) using vLLM 0.18 in enforce\_eager mode. Answers are extracted via regex matching of ``Final Answer: [A--E]'' from the response end. Accuracy is strict: the extracted letter must exactly match the correct answer.

\begin{table*}[t]
\centering
\scriptsize
\caption{Detailed experimental settings (Part I: shared configuration and offline training signals).}
\captionsetup{justification=raggedright,singlelinecheck=false,font=tiny}
\vspace{-2mm}
\label{tab:posttrain_settings_part1}
\setlength{\tabcolsep}{2pt}

\begin{tabularx}{\textwidth}{@{}p{2.8cm}X@{}}
\toprule
\textbf{Setting} & \textbf{Value} \\
\midrule

\multicolumn{2}{@{}l}{\textit{Shared Configuration (all variants)}} \\
Student model & Qwen2.5-VL-7B-Instruct (8.39B parameters) \\
Teacher model & MedMMV with GPT-5 backbone \\
LoRA rank / alpha & $r = 32$, $\alpha = 64$ \\
LoRA dropout & 0.05 \\
LoRA target modules & q\_proj, k\_proj, v\_proj, o\_proj, gate\_proj, up\_proj, down\_proj \\
Optimizer & AdamW, weight decay = 0.01 \\
Learning rate & $1 \times 10^{-5}$, cosine schedule, warmup ratio 0.15 \\
Batch size & 1 per device, gradient accumulation = 4 (effective batch = 4) \\
Max sequence length & 2,048 tokens \\
Precision & bf16 mixed precision \\
Gradient checkpointing & Enabled (non-reentrant mode) \\
Inference engine & vLLM 0.18 (enforce\_eager mode) \\
Train set & 134 samples from MedXpertQA-MM (with MedMMV teacher trajectories) \\
Test set & 100 held-out samples (eval\_all\_mc, zero overlap with train) \\

\midrule
\multicolumn{2}{@{}l}{\textit{Group A: Trace Imitation}} \\
\midrule
\textbf{Trace-SFT (raw CoT)} & \\
\quad Supervision & Teacher's first-pass CoT trace (path 1 of 3 rollouts) \\
\quad Teacher accuracy & 69.4\% on training set \\
\quad Training objective & Standard cross-entropy SFT \\
\quad Epochs & 3 \\
\quad Final train loss & 0.99 \\

\midrule
\textbf{Trace-Imitation (Teacher-Trajectory)} & \\
\quad Supervision & Final Diagnostic Reasoning section extracted from MedMMV's repaired + CU-selected trajectory \\
\quad Teacher accuracy & 77.6\% on training set \\
\quad Training objective & Standard cross-entropy SFT \\
\quad Epochs & 3 \\
\quad Final train loss & 0.99 (continued training) \\

\midrule
\multicolumn{2}{@{}l}{\textit{Group B: Preference Transfer}} \\
\midrule
\textbf{RS-Imitation (Self-Consistency)} & \\
\quad Supervision & Majority-voted trace among 3 teacher rollouts \\
\quad Teacher accuracy & 73.1\% on training set \\
\quad Selection method & Majority vote across 3 paths \\
\quad Training objective & Standard cross-entropy SFT \\
\quad Epochs & 3 \\

\midrule
\textbf{RS-Imitation (CU-Select)} & \\
\quad Supervision & CU-highest trajectory per sample \\
\quad Teacher accuracy & 72.4\% on training set \\
\quad Selection method & CU ranking (top-1) \\
\quad Training objective & Standard cross-entropy SFT \\
\quad Epochs & 3 \\

\midrule
\textbf{Pairwise-Preference (Offline)} & \\
\quad Supervision & 134 (chosen, rejected) trajectory pairs \\
\quad Pair construction & CU-top-1 vs CU-bottom-1 \\
\quad Training objective & Pairwise ranking loss $\mathcal{L} = -\log\sigma(s_\theta(x,y^+) - s_\theta(x,y^-))$ \\
\quad Score function & Normalized sequence log-probability \\
\quad Epochs & 3 \\
\quad Final train loss & 0.689 \\

\bottomrule
\end{tabularx}
\vspace{-2mm}
\end{table*}

\subsection{Trajectory Reward and Score Calibration Analysis}
\label{app:trajectory_reward}

This section provides detailed evidence that MedMMV's pipeline scores are calibrated quality signals, not arbitrary heuristics.

\begin{figure*}[htbp]
\centering
\includegraphics[width=\linewidth]{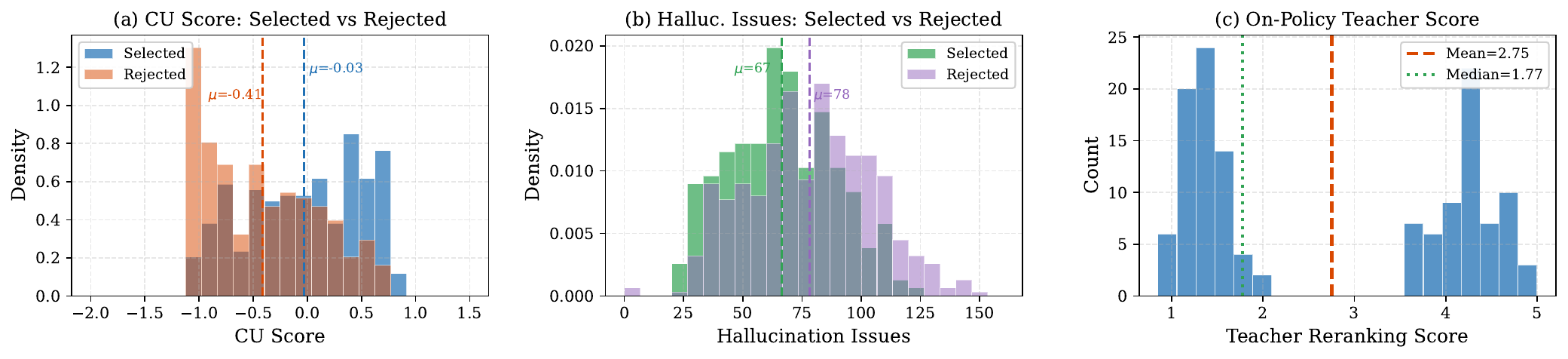}
\vspace{-6mm}
% \caption{
% Trajectory reward analysis.
% \emph{(a)} CU score distribution of selected vs.\ rejected teacher trajectories: selected paths have higher CU scores ($\mu$=$-$0.03 vs.\ $-$0.41), confirming that CU-based filtering retains higher-quality reasoning.
% \emph{(b)} Hallucination issue counts: selected trajectories exhibit fewer hallucination issues ($\mu$=67 vs.\ 78), validating that CU selection correlates with reduced fabrication.
% \emph{(c)} On-policy teacher reranking score distribution: the bimodal pattern (median=1.77, mean=2.75) reflects a clear separation between low- and high-quality student rollouts, providing a meaningful reward signal for on-policy learning.
% }
\caption{Why filtered and on-policy feedback work.
(a) Selected teacher trajectories have higher CU scores than rejected ones, confirming that CU-based filtering retains higher-quality reasoning paths.
(b) Selected trajectories also exhibit fewer hallucination issues, indicating that the filtering signal is aligned with factual quality.
(c) Teacher reranking scores on student rollouts show clear separation between low- and high-quality trajectories, providing a meaningful feedback signal for on-policy learning.}
\label{fig:trajectory_reward}
\vspace{-4mm}
\end{figure*}

\textbf{CU score separates selected from rejected trajectories.}
Figure~\ref{fig:trajectory_reward}, panel~(a) shows the CU score distribution for selected vs.\ rejected trajectories. Selected trajectories have a significantly higher mean CU score ($\mu=-0.03$ vs.\ $-0.41$), confirming that the CU scorer effectively distinguishes higher-quality reasoning paths. This gap directly translates into supervision quality for RS-Imitation and pairwise preference construction.

\textbf{Selection correlates with lower hallucination.}
Panel~(b) shows that CU-selected trajectories have fewer hallucination issues ($\mu=67$ vs.\ $78$), validating that CU captures factual grounding beyond surface-level coherence.

\textbf{On-policy teacher scores are discriminative.}
Panel~(c) shows a bimodal distribution of teacher reranking scores on student rollouts (mean~$\approx$~2.77), cleanly separating high- and low-quality student generations. This provides the discriminative reward signal that enables on-policy distillation to outperform static imitation.

\textbf{Score calibration: CU is monotonically associated with quality.}

Figure~\ref{fig:signal_calibration} provides the strongest evidence for score calibration. Binning all teacher trajectories by CU quartile, we observe:
\begin{itemize}[nosep]
    \item \emph{Accuracy}: monotonically increases from 56\% (Q1) to 88\% (Q4), with roughly equal bin sizes ($n\approx175$).
    \item \emph{Hallucination issues}: monotonically decreases from 102 (Q1) to 55 (Q4).
    \item \emph{Teacher reranking → student success}: on-policy teacher scores predict student correctness from 5\% (low tertile) to 100\% (high tertile).
\end{itemize}
These monotonic relationships confirm that CU and teacher reranking are not arbitrary selection heuristics but calibrated quality signals that meaningfully sort trajectories by both accuracy and factual grounding. This calibration is what makes CU-based filtering (RS-Imitation), pairwise preference (Pairwise-Preference (Offline)), and teacher reranking (on-policy distillation) effective supervision strategies.

\end{document}